\PassOptionsToPackage{table}{xcolor}
\documentclass{article} % For LaTeX2e
\usepackage{iclr2027_conference,times}

\usepackage{amsmath,amsfonts,bm}

\def\eqref#1{equation~\ref{#1}}
\def\1{\bm{1}}

\DeclareMathAlphabet{\mathsfit}{\encodingdefault}{\sfdefault}{m}{sl}
\SetMathAlphabet{\mathsfit}{bold}{\encodingdefault}{\sfdefault}{bx}{n}

\usepackage[utf8]{inputenc} % allow utf-8 input
\usepackage[T1]{fontenc}    % use 8-bit T1 fonts
\usepackage{graphicx}
\usepackage{hyperref}
\usepackage{url}
\usepackage{booktabs}       % professional-quality tables
\usepackage{amsfonts}       % blackboard math symbols
\usepackage{nicefrac}       % compact symbols for 1/2, etc.
\usepackage{microtype}      % microtypography
\usepackage{caption}
\usepackage{subcaption}
\usepackage{amssymb}
\usepackage[most]{tcolorbox}
\usepackage[normalem]{ulem}  % 用于删除线
\usepackage{wrapfig}
\usepackage{placeins}  % 在导言区添加
\usepackage{xcolor}         % colors
\usepackage{titletoc}
\usepackage{multirow}
\usepackage{amsmath}
\usepackage{pifont}
\definecolor{circleOrange}{RGB}{248,163,125}
\definecolor{circlePurple}{RGB}{87,63,112}
\definecolor{lightgreen}{RGB}{235, 248, 235} % 浅绿色
\definecolor{ultralightgray}{RGB}{248, 248, 248}
\definecolor{up}{RGB}{255,29,139}
\definecolor{down}{RGB}{93,208,87}
\definecolor{gray}{RGB}{236,236,236}
\definecolor{green}{RGB}{242,255,242}
\definecolor{purple}{RGB}{242,242,255}
\definecolor{yellow}{RGB}{255,254,244}
\definecolor{pink}{RGB}{255,242,242}

\definecolor{deltaRed}{RGB}{255,29,139}
\definecolor{deltaGreen}{RGB}{93,208,87}
\definecolor{resultgray}{RGB}{150,150,150}

\newcommand{\gain}[1]{\textcolor{deltaRed}{#1}}
\newcommand{\drop}[1]{\textcolor{deltaGreen}{#1}}
\newcommand{\best}[1]{\textcolor{black}{#1}}
\newcommand{\second}[1]{\textcolor{black}{\underline{#1}}}
\newcommand{\other}[1]{\textcolor{resultgray}{#1}}

\newcommand{\classlabel}[1]{\rotatebox{90}{#1}}
\newcommand{\classcolor}[1]{\textcolor{#1}{\rule{1.7ex}{1.7ex}}}

\definecolor{wallcolor}{RGB}{174,199,232}
\definecolor{floorcolor}{RGB}{152,223,138}
\definecolor{cabinetcolor}{RGB}{31,119,180}
\definecolor{bedcolor}{RGB}{255,187,120}
\definecolor{chaircolor}{RGB}{188,189,34}
\definecolor{sofacolor}{RGB}{140,86,75}
\definecolor{tablecolor}{RGB}{255,152,150}
\definecolor{doorcolor}{RGB}{214,39,40}
\definecolor{windowcolor}{RGB}{197,176,213}
\definecolor{bookshelfcolor}{RGB}{148,103,189}
\definecolor{picturecolor}{RGB}{196,156,148}
\definecolor{countercolor}{RGB}{23,190,207}
\definecolor{deskcolor}{RGB}{247,182,210}
\definecolor{curtaincolor}{RGB}{219,219,141}
\definecolor{refrigeratorcolor}{RGB}{255,127,14}
\definecolor{showercolor}{RGB}{158,218,229}
\definecolor{toiletcolor}{RGB}{44,160,44}
\definecolor{sinkcolor}{RGB}{112,128,144}
\definecolor{bathtubcolor}{RGB}{227,119,194}
\definecolor{othercolor}{RGB}{82,84,163}
\definecolor{sectiongray}{RGB}{248,248,248}

\newcommand{\cmark}{\ding{51}}
\newtcolorbox{insightbox}[2][]{
  enhanced,
  colback=blue!3,        % 背景颜色
  colframe=blue!70!black, % 边框颜色
  coltitle=white,
  fonttitle=\bfseries,
  title=#2,
  arc=2mm,
  boxrule=0.8pt,
  left=2mm,
  right=2mm,
  top=1mm,
  bottom=1mm,
  breakable,             % 允许跨页
  #1
}

\vspace{-30pt}
\title{Mantis: \underline{Ma}mba-\underline{n}ative \underline{T}uning \underline{is} Efficient for 3D Point Cloud Foundation Models}

\vspace{-30pt}
\author{%
  Zihao Guo$^{1}$ \quad
  Yiding Sun$^{1,2}$ \quad
  Jihua Zhu$^{1\thanks{Corresponding author.}}$ \quad
  Haozhe Cheng$^{3}$ \quad
  Jian Liu$^{4}$ \quad
  Ajmal Saeed Mian$^{5}$ \\
  $^{1}$Xi'an Jiaotong University, Xi'an, China,\quad
  $^{2}$Peking University, Beijing, China, \\
  $^{3}$Xi'an University of Technology, Xi'an, China, \\
  $^{4}$School of Artificial Intelligence and Robotics, Hunan University, China, \\
  $^{5}$University of Western Australia, Perth, Australia \\
  \texttt{1991002470@stu.xjtu.edu.cn, zhujh@xjtu.edu.cn}
}

\iclrfinalcopy % Uncomment for camera-ready version, but NOT for submission.
\begin{document}

\hypersetup{
    colorlinks = true,                % 去掉方框，直接显示彩色文字
    urlcolor   = [rgb]{0, 0, 1},           % 深蓝（或 navy, MidnightBlue 等）
    % linkcolor  = black,             % 内部引用颜色（可选）
    % citecolor  = black,             % 参考文献引用颜色（可选）
}

\maketitle
\vspace{-18pt}
\begin{abstract}
  \vspace{-8pt}
Pre-trained 3D point cloud foundation models (PFMs) have demonstrated strong transferability across diverse downstream tasks. However, full fine-tuning these models is computationally expensive and storage-intensive. Parameter-efficient fine-tuning (PEFT) offers a promising alternative, but existing PEFT approaches are primarily designed for Transformer-based backbones and rely on token-level prompting or feature transformation. Mamba-based backbones introduce a granularity mismatch between token-level adaptation and state-level sequence dynamics. Consequently, straightforward transfer of existing PEFT approaches to frozen Mamba backbones leads to substantial accuracy degradation and unstable optimization. To address this issue, we propose Mantis, the first Mamba-native PEFT framework for 3D PFMs. Specifically, a State-Aware Adapter (SAA) is introduced to inject lightweight task-conditioned control signals into selective state-space updates, enabling state-level adaptation while keeping the pre-trained backbone frozen. Moreover, different valid point cloud serializations are regularized by Dual-Serialization Consistency Distillation (DSCD), thereby reducing serialization-induced instability. Extensive experiments across multiple benchmarks demonstrate that Mantis achieves competitive performance while training less than 5\% of the backbone parameters. Our code is available at \href{https://github.com/gzhhhhhhh/Mantis}{\textit{Code}}.
\end{abstract}
\vspace{-8pt}
\vspace{-8pt}

\begin{figure}[htbp] 
  \centering
  \includegraphics[width=1\linewidth]{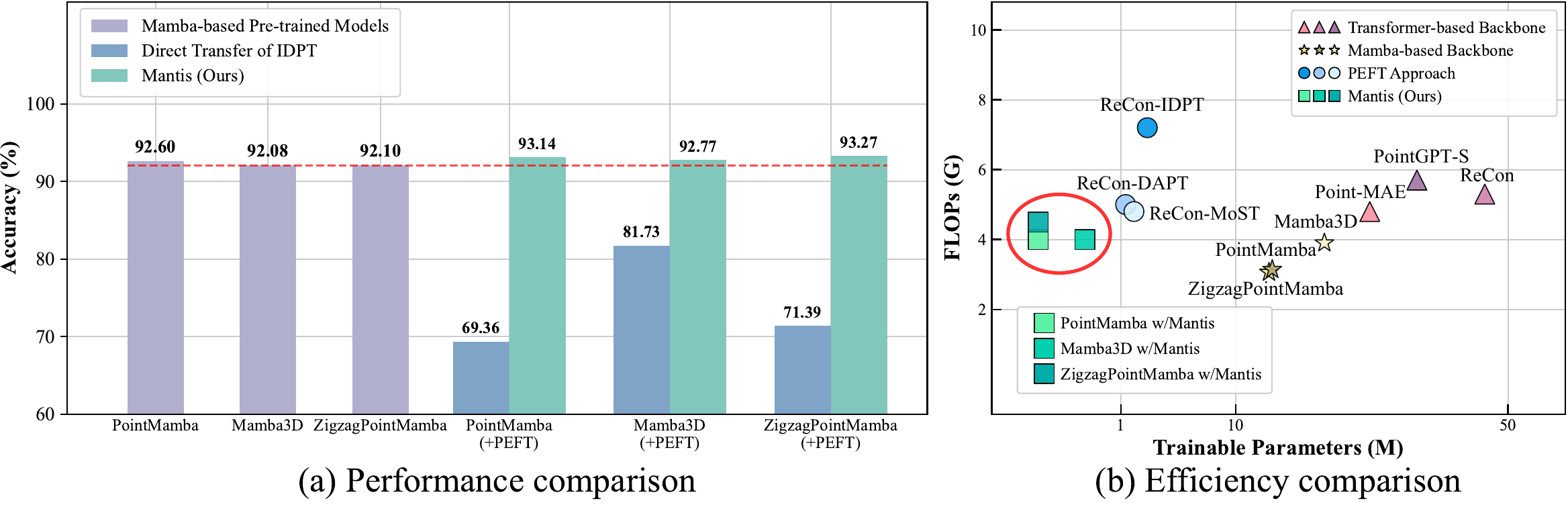} 
    \vspace{-20pt}
  \caption{Comprehensive comparisons between Mantis and several representative counterparts on OBJ\_ONLY variant of ScanObjectNN dataset. (a) Mantis achieves superior performance over existing counterparts. (b) Our Mantis achieves an ideal trade-off between trainable parameters and FLOPs.}
  \label{fig:Figure1} 
      \vspace{-10pt}
\end{figure}
\vspace{-8pt}
\section{Introduction}
\vspace{-8pt}
\label{sec:intro}
Fine-tuning pre-trained point cloud foundation models (PFMs) has been demonstrated to adapt effectively to diverse downstream tasks\citep{yu2022pointbert,pang2022pointmae,zhang2022pointm2ae,chen2023pointgpt,zha2024pointfemae}. However, the computational and storage costs associated with full fine-tuning are prohibitively high, particularly as model sizes continue to grow\citep{houlsby2019parameter,hu2022lora,tang2024point,zhou2024dapt,chen2023pointgpt,qi2023recon}. To alleviate this burden, parameter-efficient fine-tuning (PEFT) methods\citep{wang2025pointlora,liang2024pointgst,zha2025pma,han2025most,zhou2024dapt}, such as Instance-aware Dynamic Prompt Tuning (IDPT)\citep{zha2023idpt}, have emerged as promising alternatives, garnering significant attention. 
Instead of updating all model parameters directly, IDPT applies DGCNN\citep{wang2019dgcnn} to extract dynamic prompts for different instances. This approach significantly reduces the number of parameters that need to be fine-tuned, thereby lowering the computational and memory overhead. 

Follow-up research in 3D point cloud analysis can be roughly categorized into  Prompt Tuning\citep{zha2023idpt,ai2025gaprompt}, Adapters\citep{zhou2024dapt,haa2025,zha2025pma,tang2024point,liang2024pointgst} and Reparameterization\citep{han2025most,wang2025pointlora}. 
Despite these advancements, most of the prior art still relies on Transformer-native architectures\citep{pang2022pointmae,sun2026curve3d}. However, Transformer-based models incur quadratic token-length complexity and significant resource overhead, especially for large-scale point clouds and fine-grained tokenization. In parallel, recent advances in State Space Models (SSMs), particularly Mamba, offer a linear-time alternative to Transformer-based modeling\citep{guefficiently,gu2023mamba}. This advantage has already motivated a growing body of work on Mamba-based point cloud backbones, including PointMamba\citep{liang2024pointmamba} and related variants\citep{li2024mamba3d,zhang2025pcm}. 

Unfortunately, to the best of our knowledge, there is still no %genuinely 
dedicated Mamba-native PEFT framework %tailored 
designed for 3D point cloud backbones. More critically, directly transferring existing PEFT approaches to frozen Mamba-based backbones leads to substantial accuracy degradation and severe optimization instability, as illustrated in Figure~\ref{fig:Figure1} (a). The catastrophic performance drop can be attributed to two fundamental mismatches. First, Transformer-native PEFT approaches operate at the token level, whereas Mamba models rely on state-level sequence dynamics, resulting in a granularity mismatch with frozen Mamba backbones. Second, point cloud Mamba models are sensitive to serialization-dependent state propagation, %creating 
which is an inherent gap %for 
in the order-agnostic PEFT modules. This implies that indiscriminately applying existing PEFT approaches may 
%not fully exploit 
undermine 
the potential of the Mamba architecture. 
%, motivating the exploration of a Mamba-native PEFT framework.
This motivates the need for a Mamba-native PEFT framework.

In this paper, we propose Mantis, the first Mamba-native adaptation framework with two complementary components. State-Aware Adapter (SAA) integrates task-conditioned control signals into the Mamba state-space updates, bypassing token-level prompts and addressing the inherent granularity mismatch, while Dual-Serialization Consistency Distillation (DSCD) enforces consistent representations across two valid serializations of the same point cloud, regularizing order-sensitive state propagation and improving the preservation of pretrained geometric priors.

With the proposed components, Mantis achieves competitive performance across diverse downstream tasks, while training less than 5\% of the backbone parameters. Specifically, Mantis with 0.8M parameters achieves 93.29\%, 92.77\%, and 93.48\% in the three ScanObjectNN\citep{uy-scanobjectnn-iccv19} variants on Mamba3D\citep{li2024mamba3d}, surpassing full fine-tuning with 16.9M parameters by + \textbf{0.17\%, + 0.69\% and + 1.43\%}, respectively. Notably, by optimizing only lightweight task-specific modules, Mantis substantially reduces the trainable parameters and storage cost for downstream tasks, while preserving Mamba’s linear-complexity modeling paradigm, as illustrated in Figure~\ref{fig:Figure1} (b). The main contributions of this work are summarized as follows:

\begin{itemize}
    \item We reveal a fundamental gap between Transformer-native PEFT methods and Mamba-based backbones. To fill this gap, we propose Mantis, the first Mamba-native parameter-efficient fine-tuning framework for 3D point cloud analysis.
    \item Building upon our insights into the granularity mismatch, we propose \textbf{State-Aware Adapter (SAA)}, which injects task-conditioned control signals into Mamba state updates.
    \item Derived from the serialization-sensitive nature of Mamba modeling, we introduce \textbf{Dual-Serialization Consistency Distillation (DSCD)} for cross-serialization consistency and stable downstream optimization.
    \item Extensive experiments demonstrate the effectiveness and efficiency of our approach, %which casts 
    shedding light on Mamba-native PEFT for future research.
\end{itemize}
\vspace{-8pt}
\section{Related work}
\vspace{-8pt}
\label{sec:related_work}
\subsection{Point cloud foundation models}

PFMs are pre-trained on a pretext task, followed by fine-tuning for downstream applications. Architecturally, existing 3D PFMs can be divided into Transformer-based\citep{chen2023pointgpt,pang2022pointmae,zha2024pointfemae,zhao2021pointtransformer,wu2022point,wu2024pointtransformerv3} and Mamba-based\citep{liang2024pointmamba,li2024mamba3d,diao2025zigzagpointmamba,zhang2025pcm} categories. Initially proposed for natural language processing\citep{vaswani2017attention,devlin2019bert}, Transformer-based approaches leverage self-attention to capture long-range dependencies and global context in 3D point cloud analysis. For instance, Point-BERT\citep{yu2022pointbert}, Point-MAE\citep{pang2022pointmae} and Point-M2AE\citep{zhang2022pointm2ae} simplify pre-training via masked autoencoding. Multimodal integration further enriches learned representations. ReCon\citep{qi2023recon} and HyperPoint\citep{SUN2026112800} embed contrastive objectives into generative pipelines to avoid overfitting in Transformer-based approaches.

Meanwhile, Mamba-based approaches have emerged as a new paradigm for PFMs, offering efficient alternatives to Transformers by relying on SSMs. PointMamba\citep{liang2024pointmamba} pioneers point cloud modeling, providing strong global representations for downstream tasks, while Mamba3D\citep{li2024mamba3d} focuses on enhancing local features. HydraMamba\citep{qu2025hydramamba}, StruMamba3D\citep{wang2025strumamba3d} and OctMamba\citep{JIANG2026113113} employ various space-filling curves to form causal sequences for SSM. In contrast to the above methods, PCM\citep{zhang2025pcm} and CloudMamba\citep{qu2026cloudmamba} directly build structural dependencies without explicit curve parameterization. 
While these pre-trained 3D vision models with full fine-tuning have demonstrated exceptional capabilities for various downstream 3D tasks, the high computational costs and potential dilution of pre-trained knowledge motivate our exploration of efficient fine-tuning strategies. 
\subsection{Parameter-efficient fine-tuning}
As highlighted above, PFMs are iterating at an unprecedented pace in terms of structure\citep{sun2026align}, method\citep{wang2026pointrft,li2025pointdico,wu2025sonata}, and applications\citep{zhang2026pointcot}. This momentum has rendered the PEFT community equally dynamic. PEFT in 3D point cloud approaches mainly fall into prompt-based\citep{zha2023idpt,ai2025gaprompt}, adapter-based\citep{zhou2024dapt,haa2025,zha2025pma,tang2024point,liang2024pointgst} and reparameterization-based\citep{han2025most,wang2025pointlora}. Prompt-based approaches add learnable tokens to input data or attention layers to guide task-specific tuning. For example, IDPT\citep{zha2023idpt} introduces instance-aware dynamic prompts to improve robustness during downstream adaptation. Meanwhile, adapter-based pioneers like DAPT\citep{zhou2024dapt} combines dynamic adapter scaling with prompt generation to better match downstream token significance. In order to overcome the limitation in inference overhead, reparameterization-based approaches reformulate the network weights, enabling parameter-efficient adaptation. PointLoRA\citep{wang2025pointlora} brings low-rank adaptation with local token priors to fine-tuning and MoST\citep{han2025most} brings Monarch-based sparse reparameterization with local geometric priors. However, these tuning approaches are specifically designed for Transformer-based backbones, which limits their adaptability to Mamba-based models. In contrast, our Mantis, the first Mamba-native parameter-efficient fine-tuning framework for 3D point cloud analysis, fills the underexplored gap in PEFT for Mamba-based backbones and enables efficient adaptation to downstream tasks while preserving Mamba’s linear-complexity modeling.
\vspace{-8pt}
\section{Methods}
\vspace{-8pt}
\label{sec:methods}

\subsection{Overall pipeline}
\vspace{-6pt}
\begin{figure}[t] 
  \centering
  \includegraphics[width=1\linewidth]{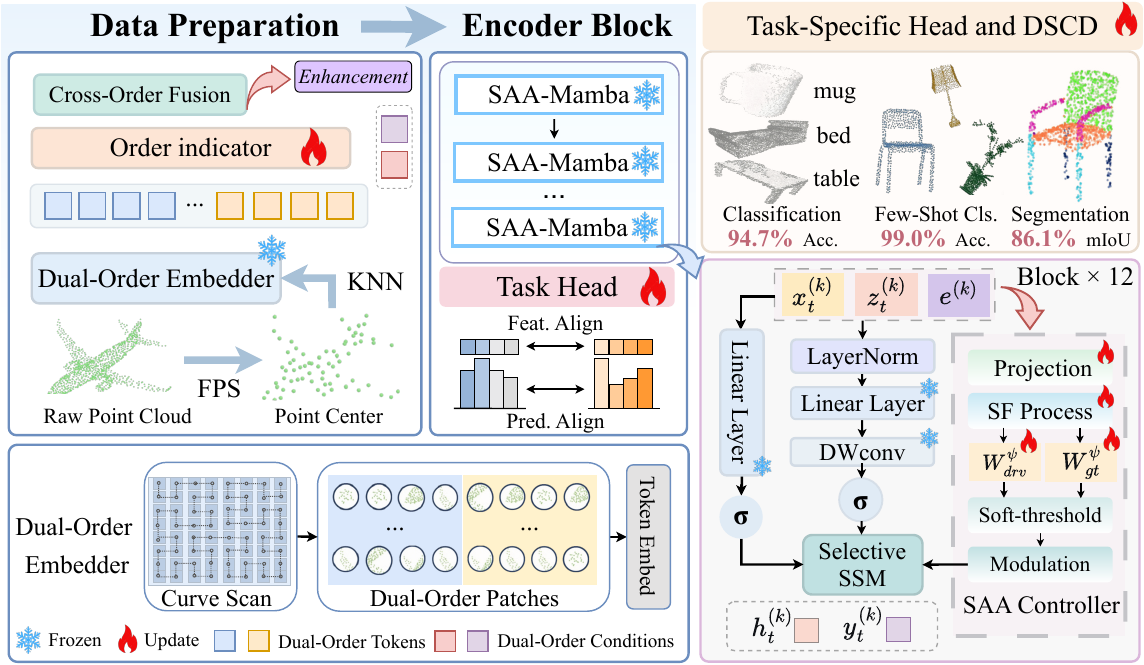} 
  \vspace{-6pt}
  \caption{\textbf{Pipeline of Mantis.} Raw point clouds are processed by Farthest Point Sampling (FPS) and K-nearest neighbors (KNN) to generate local patches. The resulting patches are serialized along two complementary space-filling curves to produce dual-order patches. These patches are processed by stacked SAA-Mamba blocks, where the selective SSM is modulated by the SAA controller. DSCD regularizes cross-serialization consistency at both feature and prediction levels during training, while only a single serialization branch is retained at inference. During training, the backbone is frozen and only newly added parameters are fine-tuned. $\sigma$ indicates SiLU activation.}
  \label{fig:Figure2} 
  \vspace{-12pt}
\end{figure}
The pipeline of our approach is shown in Figure~\ref{fig:Figure2}. Given an input point cloud $P \in \mathbb{R}^{M \times 3}$, we first utilize FPS to sample $n$ representative key points, denoted as $p \in \mathbb{R}^{n \times 3}$. The order of key points $p$ is generally random, which does not affect previous Transformer-based approaches due to their inherent order-invariance. However, for the selective SSM, its unidirectional modeling struggles with unstructured point clouds. To address this, building on the pioneer works\citep{liang2024pointmamba}, we leverage space-filling curves (\textit{e.g.}, the Hilbert curve\citep{hilbert1935stetige} and its transposed variant) for dual-order serialization, transforming the unstructured point clouds into regular sequences, denoted as $p_h$ and $p_{h'}$. We then apply the KNN algorithm to select $K$ nearest neighbors for each key point, forming $n$ local token patches $\mathcal{N}_h^{(1)} \in \mathbb{R}^{n \times K \times 3}$ and $\mathcal{N}_{h'}^{(2)} \in \mathbb{R}^{n \times K \times 3}$ with patch size $K$. To map the local patches to feature space, we employ a lightweight PointNet\citep{qi2017pointnet} to obtain serialized point tokens $E_h^{(1)} \in \mathbb{R}^{n \times d}$ and $E_{h'}^{(2)} \in \mathbb{R}^{n \times d}$.

Different from existing approaches that employ feature-wise affine transformations as order indicators\citep{liang2024pointmamba}, we model the serialization order as a conditioning signal that modulates the representation in a residual manner. Specifically, for each branch $k \in \{1,2\}$, we introduce a learnable order embedding $o^{(k)} \in \mathbb{R}^{d_o}$. The order-aware representation $e^{(k)} \in \mathbb{R}^d$ is formulated as follows:
\begin{equation}
e^{(k)} = \operatorname{MaxPool}(E^{(k)} + \mathcal{G}\!\left(E^{(k)}\right) \odot \varphi\!\left(o^{(k)}\right)),
\end{equation}
where $\mathcal{G}(\cdot)$ denotes a lightweight transformation on the input features, and $\varphi(\cdot)$ maps the order embedding to a modulation vector in the feature space. $\odot$ denotes element-wise product.
The two serialized sequences capture complementary structural cues under different traversal orders, while DSCD enforces cross-order consistency during training. The resulting representations $Z_0^{(k)} \in \mathbb{R}^{n \times d}$ are then fed into a non-hierarchical encoder composed of stacked SAA-Mamba blocks. Each SAA-Mamba block follows the general design of a Mamba layer, consisting of layer normalization, linear projection, depth-wise convolution, selective state-space modeling, and residual connection. Specifically, given the input $Z_{l-1}^{(k)}$ to the $l$-th block, we first obtain the intermediate features by:
\begin{equation}
\tilde{Z}_{l}^{(k)} = \operatorname{Linear}\left(\operatorname{LayerNorm}(Z_{l-1}^{(k)})\right), \quad
X_{l}^{(k)} = \sigma\!\left(\operatorname{DWConv}\left(\tilde{Z}_{l}^{(k)}\right)\right),
\end{equation}
where $\sigma$ indicates SiLU activation. The resulting feature matrix $X_l^{(k)}$ is then fed into the selective SSM for sequence modeling. Different from standard Mamba blocks, we introduce a \textbf{State-Aware Adapter (SAA)} controller to modulate the state transition dynamics within the selective SSM. The output of the $l$-th block is computed as:
\begin{equation}
Z_l^{(k)} = \mathcal{F}_{\text{SAA}}\!\left(X_l^{(k)}\right) + \sigma(\operatorname{Linear}(Z_{l-1}^{(k)})),
\end{equation}
where $\mathcal{F}_{\text{SAA}}(\cdot)$ denotes the selective SSM enhanced with the proposed State-Aware Adapter.
\vspace{-6pt}
\subsection{State-Aware Adapter}
\vspace{-6pt}
In the SAA-Mamba block described above, we follow the selective SSM formulation, where $x_t$, $h_t$, and $y_t$ denote the input, hidden state, and output at sequence step $t$, respectively. The state dynamics are governed by the structured transition operator $\boldsymbol{A}$ and the input-dependent selective coefficients $\boldsymbol{B}_t$, $\boldsymbol{C}_t$, and $\boldsymbol{\Delta}_t$, which control input injection, readout, and discretization, respectively. Detailed preliminaries are provided in Appendix~\ref{app:preliminaries}.
Based on this observation, we introduce a lightweight, task-conditioned
control signal to modulate the input-dependent selective coefficients,
enabling adaptation of the state evolution while preserving the structured
transition operator. In contrast to LoRA-style static weight updates\citep{hu2022lora,pmlr-v235-liu24bn}, our control signal is input-dependent and dynamically modulates the state evolution at each step. Moreover, SAA acts directly on state-space dynamics rather than inserting adaptation solely into token-level representations, providing fine-grained control over sequence evolution while keeping the backbone frozen.
At each step $t$, the SAA controller takes the current SSM input $x_t^{(k)}$, the incoming block representation $z_t^{(k)}$, and the order-aware representation $e^{(k)}$ as inputs, where $z_t^{(k)}$ denotes the $t$-th token of $Z_{l-1}^{(k)}$. Since these quantities are available before the selective scan, the control signals for all sequence positions can be generated in parallel:
\begin{equation}
\phi_t^{(k)} = \Phi\!\left(
W_x^{\phi} x_t^{(k)}
\,\|\, W_z^{\phi} z_t^{(k)}
\,\|\, W_e^{\phi} e^{(k)}
\right),
\end{equation}
where $\Phi(\cdot)$ denotes the control-feature fusion process, $\|$ denotes feature-wise concatenation, and $W_x^{\phi}, W_z^{\phi}, W_e^{\phi} \in \mathbb{R}^{d_{\phi}\times d}$ project the three pre-scan representations into a shared $d_{\phi}$-dimensional control space. We formulate control signal generation as a proximal optimization problem, which admits a closed-form soft-thresholding solution. Specifically, denote $\mathbf{q}_t^{(k)} = W_{\mathrm{drv}}^{\psi}\,\phi_t^{(k)}$, and $\boldsymbol{\lambda}_t^{(k)} = -\log\operatorname{sigmoid}\!\left(W_{\mathrm{gt}}^{\psi}\,\phi_t^{(k)}\right).$ The optimal sparse control signal is obtained by minimizing:
\begin{equation}
\mathbf{u}_t^{(k)} = \arg\min_{\mathbf{v} } \frac{1}{2}\left\|\mathbf{q}_t^{(k)} - \mathbf{v}\right\|_2^2 + \sum_i \lambda_{t,i}^{(k)} \left|v_i\right|,
\end{equation}
where $v_i$ is the $i$-th entry of the optimization variable $\mathbf{v} \in \mathbb{R}^r$. Its closed-form solution is the component-wise thresholding operator:
\begin{equation}
u_{t,i}^{(k)} =
\begin{cases}
q_{t,i}^{(k)} - \lambda_{t,i}^{(k)}, & \text{if } q_{t,i}^{(k)} > \lambda_{t,i}^{(k)}, \\[4pt]
0, & \text{if } \left|q_{t,i}^{(k)}\right| \le \lambda_{t,i}^{(k)}, \\[4pt]
q_{t,i}^{(k)} + \lambda_{t,i}^{(k)}, & \text{if } q_{t,i}^{(k)} < -\lambda_{t,i}^{(k)},
\end{cases}
\end{equation}
where $q_{t,i}^{(k)}$ and $\lambda_{t,i}^{(k)}$ denote the $i$-th element of $\mathbf{q}_t^{(k)}$ and $\boldsymbol{\lambda}_t^{(k)}$, respectively. $W_{\mathrm{drv}}^{\psi}, W_{\mathrm{gt}}^{\psi} \in \mathbb{R}^{r \times d_{\phi}}$ are learnable projection matrices, and $r \ll d$ denotes the bottleneck dimension. To achieve task-conditioned adaptation, SAA directly modulates the native
input-dependent selective coefficients
$(\boldsymbol{B}_t^{(k)}, \boldsymbol{C}_t^{(k)},
\boldsymbol{\Delta}_t^{(k)})$
via the shared sparse control signal $\mathbf{u}_t^{(k)}$:
\begin{equation}
\widetilde{\boldsymbol{O}}_t^{(k)}
=
\boldsymbol{O}_t^{(k)}
+
\mathcal{R}_{O}\!\left(
\mathbf{U}_{O}
\operatorname{diag}\!\left(\mathbf{u}_t^{(k)}\right)
\mathbf{V}_{O}
\right),
\qquad
O \in \{B,C,\Delta\}.
\end{equation}
where $\mathbf{U}_{O}$ and $\mathbf{V}_{O}$ are coefficient-specific low-rank modulation matrices, and $\mathcal{R}_{O}(\cdot)$ maps each factorized correction to the native coefficient layout. The structured transition operator $\boldsymbol{A}$ remains frozen. Since $\mathbf{u}_t^{(k)}$ depends only on pre-scan representations, all modulated coefficients are generated before recurrence and supplied to the backbone's native fused selective-scan operator. The structured discretization performed within the selective SSM can be written as:
\begin{equation}
\widehat{\boldsymbol{A}}_t^{(k)}
=
\exp\!\left(
\boldsymbol{A}\,\widetilde{\boldsymbol{\Delta}}_t^{(k)}
\right),
\quad
\widehat{\boldsymbol{B}}_t^{(k)}
=
\boldsymbol{A}^{-1}
\left(
\exp\!\left(
\boldsymbol{A}\,\widetilde{\boldsymbol{\Delta}}_t^{(k)}
\right)
-
\boldsymbol{I}
\right)
\widetilde{\boldsymbol{B}}_t^{(k)}.
\end{equation}
Eq.~(8) describes the corresponding structured discretization mathematically. In implementation, these quantities are evaluated through the native factorized selective-scan computation without explicitly materializing dense transition matrices or general matrix exponentials.
As a consequence, SAA formulates the selective SSM as a conditionally controlled dynamical system while preserving its native recurrence. Importantly, all SAA coefficients are computed in parallel before a single native selective-scan call per SAA-Mamba block and serialization branch; no auxiliary recurrent pass or recurrent approximation is introduced. Appendix~\ref{app:saa_kernel} further analyzes the induced selective-kernel perturbation and hidden-state deviation.
\subsection{Dual-Serialization Consistency Distillation}

Although serialization enables Mamba to process unordered point clouds as sequential inputs, the resulting hidden-state propagation is inherently sensitive to the specific traversal order. Consequently, different valid serializations of the same point cloud may induce inconsistent feature representations and unstable predictions, which hinders reliable adaptation under frozen Mamba backbones. We provide a transfer-kernel analysis of this serialization sensitivity in Appendix~\ref{app:selective_ss}.

\begin{wrapfigure}{r}{0.6\textwidth}
\vspace{-8pt}
\centering
\includegraphics[width=\linewidth]{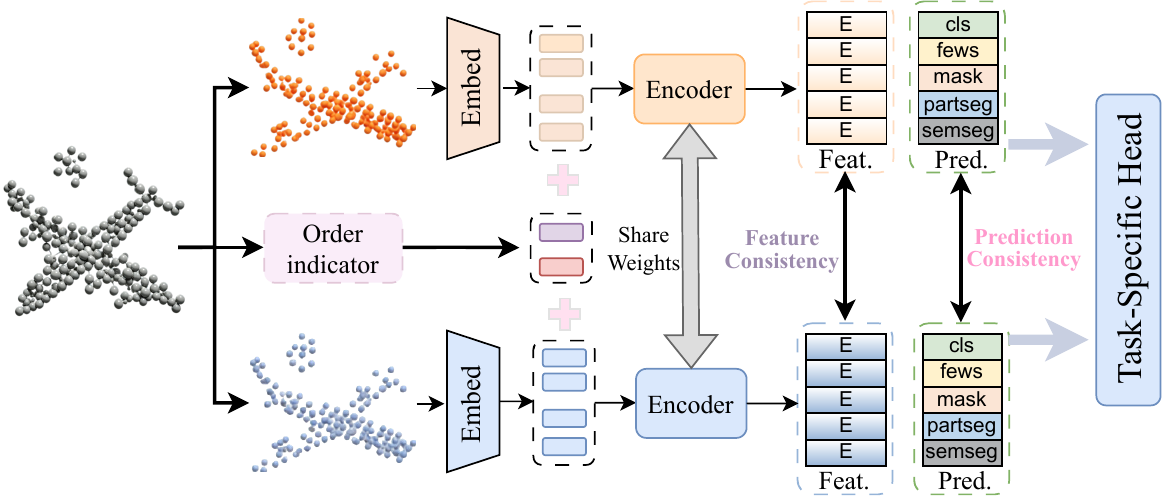}
\vspace{-16pt}
\caption{\textbf{Illustration of Dual-Serialization Consistency Distillation.} During training, DSCD enforces cross-serialization consistency at both feature and prediction levels.}
\label{fig:Figure36}
\vspace{-14pt}
\end{wrapfigure}
\label{sec:dscd}

To alleviate this issue, we propose \textbf{Dual-Serialization Consistency Distillation (DSCD)}, which enforces cross-order consistency by aligning feature representations and predictive distributions across two serialization views, thereby reducing serialization-induced discrepancies, as illustrated in Figure~\ref{fig:Figure36}. Unlike generic augmentation consistency, the two views preserve exactly the same underlying point geometry but induce distinct traversal-dependent state propagation patterns in Mamba. Built upon the dual-serialization pipeline described above, DSCD regularizes the two serialization branches at the final encoder layer. The final-layer output of branch $k$ is denoted by $Z_{L}^{(k)}$, and the feature alignment loss is formulated as:

\begin{equation}
z^{(k)} = g_{\varphi}\!\left(\operatorname{AvgPool}(Z_{L}^{(k)})\right), \qquad
\mathcal{L}_{\mathrm{feat}}
=
\left\|
\frac{z^{(1)}}{\|z^{(1)}\|_2}
-
\frac{z^{(2)}}{\|z^{(2)}\|_2}
\right\|_2^2,
\end{equation}
where $g_{\varphi}$ is a branch-shared projection mapping. While $\mathcal{L}_{\mathrm{feat}}$ enforces representation-level consistency across different serialization orders, stable downstream optimization further requires alignment at the decision level. The prediction-level consistency term is defined as:
\begin{equation}
\pi^{(k)}=\operatorname{Softmax}\!\left(\ell^{(k)}/\tau\right),
\qquad 
\mathcal{L}_{\mathrm{pred}}
=
\frac{1}{2}\,\mathrm{KL}\!\left(\pi^{(1)} \,\|\, \pi^{(2)}\right)
+
\frac{1}{2}\,\mathrm{KL}\!\left(\pi^{(2)} \,\|\, \pi^{(1)}\right),
\end{equation}
where $\tau$ is the temperature hyperparameter, and $\pi^{(k)}$ denotes the softened predictive distribution of branch $k$. Accordingly, the training objective can be defined as follows:
\begin{equation}
\mathcal{L}
=
\mathcal{L}_{\mathrm{task}}
+
\alpha\,\mathcal{L}_{\mathrm{feat}}
+
\beta\,\mathcal{L}_{\mathrm{pred}},
\end{equation}
where $\alpha$ and $\beta$ are coefficients for feature consistency and prediction consistency, respectively. By reducing cross-serialization discrepancies at both the representation and prediction levels, DSCD improves cross-serialization consistency and stabilizes downstream optimization.

\section{Experiments}
\label{sec:experiments}
\begin{table}[t]
\centering
\Large
\caption{Classification on three variants of ScanObjectNN\citep{uy-scanobjectnn-iccv19} and ModelNet40\citep{wu20153d}, including the number of trainable parameters, inference FLOPs and overall accuracy (OA). \#TP (M) represents the trainable parameters. Attn. denotes attention-based (Transformer) and SSM denotes state-space model (Mamba). $^\dagger$ indicates that using simple rotational augmentation\citep{dongautoencoders} for training. \textcolor{circleOrange}{$\circ$} and \textcolor{circlePurple}{$\circ$} denote methods that leverage cross-modal and single-modal information, respectively. ModelNet40 results report the overall accuracy from 1024 points without voting. Our results are averaged over three random seeds. * denotes results reproduced from the public source code. The best two results are bolded and underlined respectively. Same as follows.}
\label{tab:classification}
\vspace{-6pt}
\resizebox{\textwidth}{!}{%
\renewcommand{\arraystretch}{1.0}%
\begin{tabular}{l l l c c c c c c}
\toprule
\multirow{2}{*}{Methods} & 
\multirow{2}{*}{Reference} & 
\multirow{2}{*}{Backbone} & 
\multirow{2}{*}{\#TP (M)$\downarrow$} & 
\multirow{2}{*}{FLOPs (G)} & 
\multicolumn{3}{c}{ScanObjectNN$\uparrow$} & 
\multicolumn{1}{c}{ModelNet40$\uparrow$} \\
\cmidrule(lr){6-8} \cmidrule(lr){9-9}
& & & & & OBJ\_BG & OBJ\_ONLY & PB\_T50\_RS & OA (\%) \\
\midrule
\rowcolor{ultralightgray}\multicolumn{9}{c}{\textit{Supervised Learning Only}} \\
\midrule
{\color{circlePurple}$\circ$} PointNet\citep{qi2017pointnet}         & CVPR17    & \multicolumn{1}{c}{-}           & 3.5  & 0.5  & 73.3  & 79.2  & 68.0  & 89.2\\
{\color{circlePurple}$\circ$} PointNet++\citep{qi2017pointnet2}       & NeurIPS17 & \multicolumn{1}{c}{-}           & 1.5  & 1.7  & 82.3  & 84.3  & 77.9 & 90.7\\
{\color{circlePurple}$\circ$} PointCNN\citep{li2018pointcnn}         & NeurIPS18 & \multicolumn{1}{c}{-}           & 0.6  & 0.9  & 86.1  & 85.5  & 78.5 & 92.2\\
{\color{circlePurple}$\circ$} DGCNN\citep{wang2019dgcnn}            & TOG19     & \multicolumn{1}{c}{-}           & 1.8  & 2.4  & 82.8  & 86.2  & 78.1 & 92.9\\
\midrule
\rowcolor{ultralightgray}\multicolumn{9}{c}{\textit{Self-Supervised Learning (Full Fine-tuning)}} \\
\midrule
{\color{circlePurple}$\circ$} Point-BERT\citep{yu2022pointbert}       & CVPR22    & \multicolumn{1}{c}{Attn.} & 22.1 & 4.8  & 87.43 & 88.12 & 83.07 & 93.2\\
{\color{circlePurple}$\circ$} Point-M2AE\citep{zhang2022pointm2ae}       & NeurIPS22 & \multicolumn{1}{c}{Attn.} & 15.3 & 3.6  & 91.22 & 88.81 & 86.43 & 94.0\\
{\color{circlePurple}$\circ$} PointGPT-S$^\dagger$\citep{chen2023pointgpt}       & NeurIPS23 & \multicolumn{1}{c}{Attn.} & 29.2 & 5.7  & 93.39 & 92.43 & 89.17 & 93.3\\
{\color{circleOrange}$\circ$} ACT$^\dagger$\citep{dongautoencoders}              & ICLR23    & \multicolumn{1}{c}{Attn.} & 22.1 & 4.8  & 93.29 & 91.91 & 88.21 & 93.6\\
{\color{circleOrange}$\circ$} Joint-MAE\citep{guo2023joint}        & IJCAI23   & \multicolumn{1}{c}{Attn.} & -    & -    & 90.94 & 88.86 & 86.07 & 94.0 \\
{\color{circleOrange}$\circ$} I2P-MAE$^\dagger$\citep{zhang2023learning}          & CVPR23    & \multicolumn{1}{c}{Attn.} & 15.3 & -    & 94.15 & 91.57 & 90.11 & 93.4\\
{\color{circlePurple}$\circ$} PCM\citep{zhang2025pcm}       & AAAI25 & \multicolumn{1}{c}{SSM} & 34.2 & - & - & - & 88.10 & 93.4\\
{\color{circlePurple}$\circ$} SI-Mamba\citep{bahri2025spectral}       & CVPR25 & \multicolumn{1}{c}{SSM} & 12.3 & 3.6 & 92.25 & 91.39 & 87.30 & 92.7\\
\midrule
\rowcolor{ultralightgray}\multicolumn{9}{c}{\textit{Self-Supervised Learning (Parameter-Efficient Fine-Tuning)}} \\
\midrule
{\color{circlePurple}$\circ$} Point-MAE\citep{pang2022pointmae}        & ECCV22    & \multicolumn{1}{c}{Attn.} & 22.1 (100\%) & 4.8  & 90.02 & 88.29 & 85.18 & 93.8\\
\  \ + IDPT\citep{zha2023idpt} & ICCV23 & \multicolumn{1}{c}{Attn.} & 1.7 (7.7\%) & 7.2 & 91.22 ({\color{up}+ 1.20}) & 90.02 ({\color{up}+ 1.73}) & 84.94 ({\color{down}- 0.24}) & 93.3 ({\color{down}- 0.5})\\
\  \ + PointLoRA\citep{wang2025pointlora} & CVPR25 & \multicolumn{1}{c}{Attn.} & 0.8 (3.6\%) & - & 90.71 ({\color{up}+ 0.69}) & 89.33 ({\color{up}+ 1.04}) & 85.53 ({\color{up}+ 0.35}) & 93.3 ({\color{down}- 0.5})\\
\  \ + PoinTriE\textsuperscript{*}\citep{sun2026tri} & ECCV26 & \multicolumn{1}{c}{Attn.} & 1.1 (5.0\%) & 4.2 & 91.30 ({\color{up}+ 1.28}) & 90.13 ({\color{up}+ 1.84}) & 85.71 ({\color{up}+ 0.53}) & 93.7 ({\color{down}- 0.1})\\\midrule
{\color{circleOrange}$\circ$} ReCon$^\dagger$\citep{qi2023recon}           & ICML23    & \multicolumn{1}{c}{Attn.} & 43.6 (100\%) & 5.3  & \textbf{95.18} & \textbf{93.63} & 90.63 & \underline{94.1}\\
\  \ + DAPT\citep{zhou2024dapt} & CVPR24 & \multicolumn{1}{c}{Attn.} & 1.1 (2.5\%) & 5.0 & 94.32 ({\color{down}- 0.86}) & 92.43 ({\color{down}- 1.20}) & 89.38 ({\color{down}- 1.25}) & 93.5 ({\color{down}- 0.6})\\
\  \ + GAPrompt\citep{ai2025gaprompt} & ICML25 & \multicolumn{1}{c}{Attn.} & 0.6 (1.4\%) & 5.0 & 94.49 ({\color{down}- 0.69}) & 92.60 ({\color{down}- 1.03}) & 89.76 ({\color{down}- 0.87}) & 94.0 ({\color{down}- 0.1})\\
\  \ + PoinTriE\textsuperscript{*}\citep{sun2026tri} & ECCV26 & \multicolumn{1}{c}{Attn.} & 0.8 (1.8\%) & 4.8 & 94.91 ({\color{down}- 0.27}) & 93.19 ({\color{down}- 0.44}) & 90.11 ({\color{down}- 0.52}) & \underline{94.1} ({\color{up}+ 0.0})\\
\midrule
{\color{circlePurple}$\circ$} PointMamba$^\dagger$\citep{liang2024pointmamba}       & NeurIPS24 & \multicolumn{1}{c}{SSM} & 12.3 (100\%) & 3.1 & 94.32 & 92.60 & 89.31 & 93.6\\
\  \ + PMA\textsuperscript{*}\citep{zha2025pma}       & CVPR25 & \multicolumn{1}{c}{SSM} & 1.1 (8.9\%) & 5.2 & 93.53 ({\color{down}- 0.79}) & 91.32 ({\color{down}- 1.28}) & 86.71 ({\color{down}- 2.60}) & 93.6 ({\color{up}+ 0.0})\\
\  \ + PoinTriE\textsuperscript{*}\citep{sun2026tri} & ECCV26 & \multicolumn{1}{c}{SSM} & 0.8 (6.5\%) & 4.6 & 92.81 ({\color{down}- 1.51}) & 91.35 ({\color{down}- 1.25}) & 86.97 ({\color{down}- 2.34}) & 93.7 ({\color{up}+ 0.1})\\
\rowcolor{lightgreen}\  \ + Mantis$^\dagger$ & \textbf{This Paper} & \multicolumn{1}{c}{SSM} & 0.6 (4.9\%) & 4.0 & \underline{94.65} ({\color{up}+ 0.33}) & 93.14 ({\color{up}+ 0.54}) & 89.96 ({\color{up}+ 0.65})& 93.7 ({\color{up}+ 0.1 })\\
\midrule
{\color{circlePurple}$\circ$} Mamba3D\citep{li2024mamba3d}          & ACM MM24 & \multicolumn{1}{c}{SSM} & 16.9 (100\%) & 3.9 & 93.12 & 92.08 & \underline{92.05} & \textbf{94.7} \\
\  \ + PMA\textsuperscript{*}\citep{zha2025pma}       & CVPR25& \multicolumn{1}{c}{SSM} & 1.3 (7.7\%) & 4.7 & 91.97 ({\color{down}- 1.15}) & 89.64 ({\color{down}- 2.44}) & 88.70 ({\color{down}- 3.35}) & 94.2 ({\color{down}- 0.5}) \\
\  \ + PoinTriE\textsuperscript{*}\citep{sun2026tri} & ECCV26 & \multicolumn{1}{c}{SSM} & 1.0 (5.9\%) & 4.6 & 92.11 ({\color{down}- 1.01}) & 90.03 ({\color{down}- 2.05}) & 88.82 ({\color{down}- 3.23}) & 94.4 ({\color{down}- 0.3})\\
\rowcolor{lightgreen}\  \ + Mantis & \textbf{This Paper} & \multicolumn{1}{c}{SSM} & 0.8 (4.7\%) & 4.0 & 93.29 ({\color{up}+ 0.17}) & 92.77 ({\color{up}+ 0.69}) & \textbf{93.48} ({\color{up}+ 1.43}) & \textbf{94.7} ({\color{up}+ 0.0})\\
\midrule
{\color{circlePurple}$\circ$} 
ZigzagPointMamba\citep{diao2025zigzagpointmamba}       & NeurIPS25 & \multicolumn{1}{c}{SSM} & 12.3 (100\%) & 3.1 & 94.15 & 92.10 & 88.65 & 93.2\\
\  \ + PMA\textsuperscript{*}\citep{zha2025pma}       & CVPR25& \multicolumn{1}{c}{SSM} & 1.1 (8.9\%) & 5.7 & 93.10 ({\color{down}- 1.05}) & 89.53 ({\color{down}- 2.57}) & 85.18 ({\color{down}- 3.47}) & 92.8 ({\color{down}- 0.4})\\
\  \ + PoinTriE\textsuperscript{*}\citep{sun2026tri} & ECCV26 & \multicolumn{1}{c}{SSM} & 0.8 (6.5\%) & 5.2 & 93.37 ({\color{down}- 0.78}) & 90.17 ({\color{down}- 1.93}) & 85.92 ({\color{down}- 2.73}) & 93.1 ({\color{down}- 0.1})\\
\rowcolor{lightgreen}\  \ + Mantis & \textbf{This Paper} & \multicolumn{1}{c}{SSM} & 0.6 (4.9\%) & 4.5 & 94.02 ({\color{down}- 0.13}) & \underline{93.27} ({\color{up}+ 1.17}) & 90.11 ({\color{up}+ 1.46}) & 93.7 ({\color{up}+ 0.5})\\
\bottomrule
\end{tabular}%
}
\vspace{-12pt}
\end{table}

We evaluate Mantis on multiple downstream tasks, including object classification, part and semantic segmentation. We utilize three pre-trained models (PointMamba\citep{liang2024pointmamba}, Mamba3D\citep{li2024mamba3d}, and ZigzagPointMamba\citep{diao2025zigzagpointmamba}) as our baselines. For a fair comparison, we follow the default fine-tuning settings of each baseline. The hyperparameters are set as $\alpha = 100$, $\beta = 0.05$, and $\tau = 1$. Additional analyses of $\alpha$ and $\beta$ and training details can be found in Appendix~\ref{sec:app_more}.
\vspace{-12pt}
\subsection{Object classification}
\textbf{Real-World Object Classification.} The ScanObjectNN\citep{uy-scanobjectnn-iccv19} dataset is a highly challenging 3D dataset covering $\sim$15K real-world objects across 15 categories. As shown in Table~\ref{tab:classification}, we conduct experiments on three variants of ScanObjectNN (\emph{i.e.}, OBJ\_BG, OBJ\_ONLY, and PB\_T50\_RS), each with increasing complexity. Mantis surpasses full fine-tuning in most cases with only about 5\% of the parameters. Specifically, Mantis surpasses the full fine-tuning of PointMamba, Mamba3D, and ZigzagPointMamba by 0.65\%, 1.43\%, 1.46\% on PB\_T50\_RS split, respectively. Furthermore, the t-SNE feature visualizations\citep{JMLR:v9:vandermaaten08a} of existing methods and Mantis are presented in Figure~\ref{fig:visal} (a). This suggests that Mantis is especially effective in improving robustness under noisy and order-sensitive real-world point cloud distributions. Notably, the more pronounced performance improvements on the most challenging split indicate that dynamic state-aware adaptation and cross-serialization consistency become increasingly beneficial when serialization-induced instability is amplified. A detailed analysis of this phenomenon is provided in Appendix~\ref{sec:app_hardest}.

\textbf{Synthetic Object Classification.} The ModelNet40\citep{wu20153d} dataset contains a total of 12,311 3D CAD models across 40 categories. Due to the computational cost of the voting strategy\citep{liu2019relation}, we report overall accuracy without voting. As shown in Table~\ref{tab:classification}, compared to full fine-tuning, Mantis substantially reduces resource requirements while achieving comparable performance.

\vspace{-24pt}
\subsection{3D Dense Prediction Task}

For dense prediction tasks, including part and semantic segmentation, we adopt the standard task-specific heads of the corresponding baselines for fair comparison. To further establish the practical efficiency of Mantis, we additionally report broader resource metrics in Appendix~\ref{app:efficiency}.  

\begin{wrapfigure}{r}{0.5\textwidth}
\vspace{-12pt}
\centering
\includegraphics[width=1\linewidth]{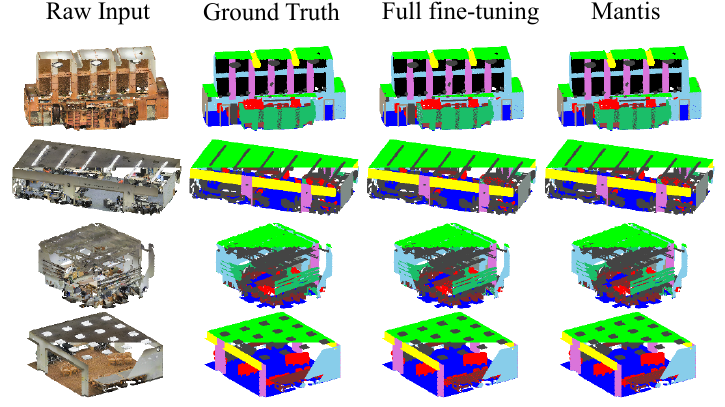}
\vspace{-20pt}
\caption{Qualitative comparison on the S3DIS.}
\label{fig:s3dis}
\vspace{-16pt}
\end{wrapfigure}

\textbf{Part Segmentation.} As shown in Table~\ref{tab:segmentation}, in this challenging fine-grained understanding task\citep{shapenetpart}, our approach still achieves competitive results on instance-level mIoU (Inst.\ mIoU) and consistent improvements on class-level mIoU (Cls.\ mIoU) compared with full fine-tuning. Qualitatively, as shown in Figure~\ref{fig:visal} (b), Mantis produces more accurate and coherent part boundaries than full fine-tuning, especially for small or structurally complex parts, thereby preserving fine-grained geometric structures.

\begin{wrapfigure}{r}{0.5\textwidth}
\vspace{-12pt}
\centering
\includegraphics[width=1\linewidth]{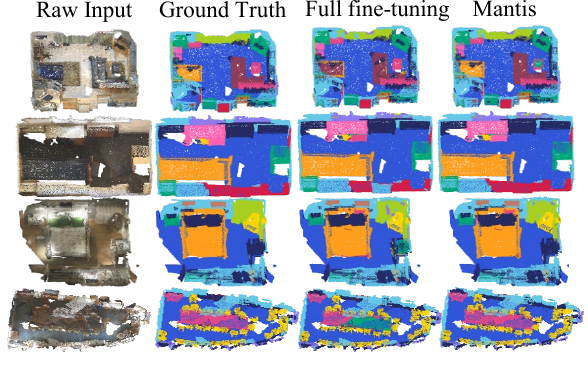}
\vspace{-24pt}
\caption{Qualitative results on the ScanNet V2.}
\label{fig:scannet}
\vspace{-24pt}
\end{wrapfigure}

\begin{figure}[t] 
  \centering
  \includegraphics[width=1\linewidth]{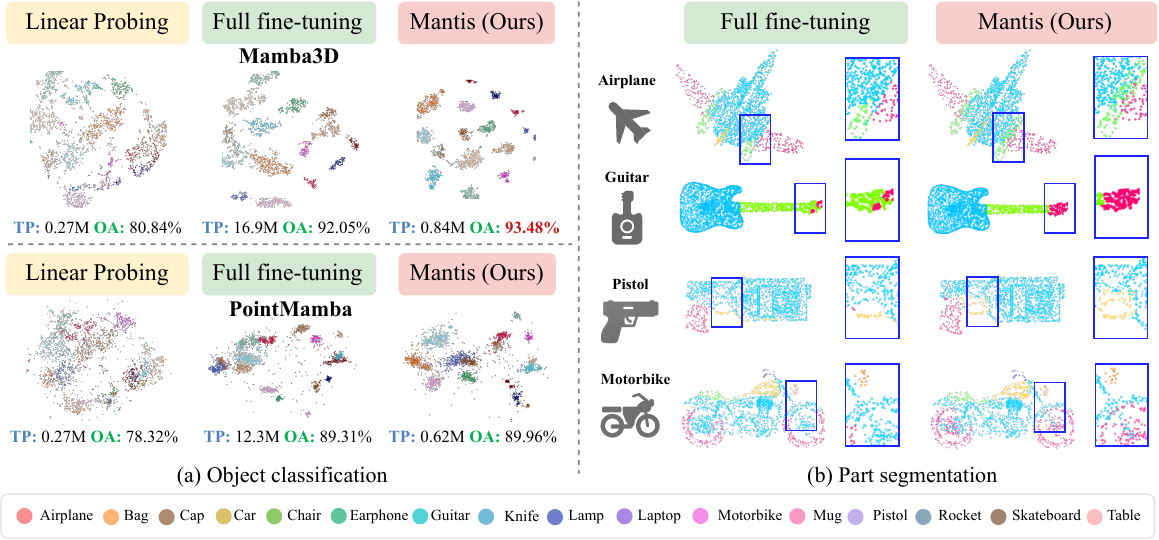} 
  \vspace{-18pt}
  \caption{\textbf{Qualitative analysis results for object classification and part segmentation.} (a) The t-SNE visualization results on the PB\_T50\_RS variant with different fine-tuning schemes. (b) Comparison of full fine-tuning and Mantis on part segmentation with Mamba3D\citep{li2024mamba3d}.}
  \label{fig:visal} 
  \vspace{-18pt}
\end{figure}

\textbf{Semantic Segmentation.} As shown in Table~\ref{tab:s3dis_scannet}, we report mean class accuracy (mAcc) and mean IoU (mIoU) on S3DIS Area 5~\citep{Armeni_2016_CVPR}, and voxel-level mIoU (VmIoU) and point-level mIoU (PmIoU) on ScanNet V2~\citep{dai2017scannet}. Due to their pretraining primarily on object-level datasets, these backbones struggle to transfer effectively to large-scale indoor scene segmentation\citep{liang2024pointgst}. In contrast, Mantis adapts state dynamics while reducing serialization variation, bridging the gap between fully and efficiently fine-tuning strategies. With Mamba3D as the backbone on S3DIS, Mantis achieves 68.0\% mAcc and 58.1\% mIoU with only 0.7M trainable backbone parameters, closely matching full fine-tuning at 68.2\% and 58.4\%. As further shown in Figure~\ref{fig:s3dis} and Figure~\ref{fig:scannet}, Mantis produces reliable dense predictions in complex indoor scenes. Overall, Mantis remains competitive with full fine-tuning on these challenging tasks, an uncommon achievement in PEFT.

\begin{table}[t]
\centering
\small

\caption{
Results on large-scale semantic segmentation on
S3DIS~\citep{Armeni_2016_CVPR} and ScanNet V2~\citep{dai2017scannet}, and part segmentation on
ShapeNetPart~\citep{shapenetpart}. Only trainable backbone parameters (Param.) are reported, as the increase mainly comes from the shared task head.
}
\label{tab:segmentation_results}

% =========================================================
% Right: ShapeNetPart
% =========================================================
\begin{minipage}[t]{0.47\linewidth}
\centering
\vspace{-6pt}
\subcaption{Part segmentation results.}
\vspace{-4pt}
\label{tab:segmentation}

\resizebox{\linewidth}{!}{%
\renewcommand{\arraystretch}{1}%
\setlength{\tabcolsep}{3pt}
\begin{tabular}{lccc}
\toprule
Methods & Param. & Inst.mIoU & Cls.mIoU \\
\midrule

\rowcolor{ultralightgray}
\multicolumn{4}{c}{
\textit{Self-Supervised Learning (Full Fine-tuning)}
} \\
\midrule

Point-BERT~\citep{yu2022pointbert}
& 21.9 & 85.6 & 84.1 \\

Point-MAE~\citep{pang2022pointmae}
& 21.9 & \underline{86.1} & 84.1 \\

ACT\citep{dongautoencoders}   & 27.1 & \underline{86.1} & \textbf{84.7} \\

ReCon~\citep{qi2023recon}
& 43.3 & \textbf{86.4} & \underline{84.5} \\
\midrule

\rowcolor{ultralightgray}
\multicolumn{4}{c}{
\textit{Self-Supervised Learning (Parameter-Efficient Fine-Tuning)}
} \\
\midrule

PointMamba
& 12.2 & 85.3 & 82.6 \\

\quad + PMA
& 1.3
& 84.2 ({\color{down}- 1.1})
& 82.3 ({\color{down}- 0.3}) \\

\rowcolor{lightgreen}
\quad \textbf{+ Mantis}
& 0.6
& 85.1 ({\color{down}- 0.2})
& 82.9 ({\color{up}+ 0.3}) \\
\midrule
Mamba3D
& 15.5 & 85.6 & 83.6 \\

\quad + PMA
& 1.8
& 83.7 ({\color{down}- 1.9})
& 82.4 ({\color{down}- 1.2}) \\

\rowcolor{lightgreen}
\quad \textbf{+ Mantis}
& 0.7
& \underline{86.1} ({\color{up}+ 0.5})
& 84.1 ({\color{up}+ 0.5}) \\
\midrule
ZigzagPointMamba
& 12.2 & 85.8 & 84.2 \\

\quad + PMA
& 1.3
& 84.1 ({\color{down}- 1.7})
& 82.0 ({\color{down}- 2.2}) \\

\rowcolor{lightgreen}
\quad \textbf{+ Mantis}
& 0.6
& 85.9 ({\color{up}+ 0.1})
& \underline{84.5} ({\color{up}+ 0.3}) \\

\bottomrule
\end{tabular}%
}

\end{minipage}
\hfill
% ========================================================
% Left: S3DIS Area 5
% =========================================================
\begin{minipage}[t]{0.52\linewidth}

\centering
\vspace{-6pt}
\subcaption{Indoor scene semantic segmentation results.}
\vspace{-4pt}
\label{tab:s3dis_scannet}

\resizebox{\linewidth}{!}{%
\renewcommand{\arraystretch}{1.01}%
\setlength{\tabcolsep}{4pt}
\begin{tabular}{lc|cc|cc}
\toprule

\multirow{2}{*}{Method}
& \multirow{2}{*}{Param.}
& \multicolumn{2}{c|}{S3DIS Area 5}
& \multicolumn{2}{c}{ScanNet V2} \\

\cline{3-4}
\cline{5-6}

&
& \raisebox{-2pt}{mAcc} & \raisebox{-2pt}{mIoU}
& \raisebox{-2pt}{VmIoU} & \raisebox{-2pt}{PmIoU} \\

\midrule

\rowcolor{ultralightgray}
\multicolumn{6}{c}{
\textit{Self-Supervised Learning (Full Fine-tuning)}
} \\

\midrule
Point-BERT~\citep{yu2022pointbert}
& 21.8
& 69.7 & 60.5
& 49.9 & 49.6 \\

ReCon~\citep{qi2023recon}
& 21.8
& 69.3 & 60.4
& 50.8 & 50.5 \\ 
\midrule

\rowcolor{ultralightgray}
\multicolumn{6}{c}{
\textit{Self-Supervised Learning (Parameter-Efficient Fine-Tuning)}
} \\
\midrule

Point-MAE
& 21.8
& 69.9 & 60.8
& 51.2 & 50.8 \\

\quad + PointLoRA
& 2.2
& 64.8
& 52.6
& 38.1 & 37.2 \\

\quad + PMA
& 2.4
& 65.0
& 53.2
& 39.3 & 37.9 \\

\quad + PoinTriE
& 1.9
& 66.1
& 54.9
& 43.4 & 42.1 \\
\midrule

PointMamba
& 12.2
& 70.1 & 60.9
& 50.3 & 50.1 \\

\quad + PMA
& 1.2
& 63.8
& 54.6
& 42.7 & 41.6 \\

\rowcolor{lightgreen}
\quad \textbf{+ Mantis}
& 0.6
& 69.3
& 59.8
& 49.8 & 49.5 \\

\midrule

Mamba3D
& 15.4
& 68.2 & 58.4
& 51.3 & 50.7 \\

\quad + PMA
& 1.8
& 64.1
& 54.6
& 43.1 & 42.3 \\

\rowcolor{lightgreen}
\quad \textbf{+ Mantis}
& 0.7
& 68.0
& 58.1
& 50.4 & 50.0 \\

\bottomrule
\end{tabular}%
}

\end{minipage}

\end{table}

\begin{table}[t]
\vspace{-6pt}
\centering
\small
\caption{Ablation study on the setting of Mantis, including different components, inserted layers and PEFT methods. The details of the table are described in Sec.~\ref{subsec:ablation}.}
\label{tab:ablation}
\begin{minipage}[t]{0.34\textwidth}
\centering
\vspace{-6pt}
\subcaption{The effect of each component.}
\label{tab:ablation_components}
{%
\resizebox{\linewidth}{!}{%
\renewcommand{\arraystretch}{1.09}%
\begin{tabular}{ccccc}
\toprule
\multirow{2}{*}{SAA} & \multicolumn{2}{c}{DSCD} & \multirow{2}{*}{\#TP (M)} & \multirow{2}{*}{PB\_T50\_RS} \\
\cmidrule(lr){2-3}
& $\mathcal{L}_{\mathrm{feat}}$ & $\mathcal{L}_{\mathrm{pred}}$ & & \\
\midrule
\rowcolor{ultralightgray}\multicolumn{5}{c}{\textit{Vanilla Protocol}} \\
\midrule
\multicolumn{3}{c}{Full fine-tuning} & 16.9 & 92.05 \\
\multicolumn{3}{c}{Linear Probing} & 0.27 & 80.84 \\
\midrule

       & \cmark &        & 0.38 & 84.12 \\
       &        & \cmark & 0.38 & 83.95 \\
       & \cmark & \cmark & 0.38 & 84.53 \\
\cmark &        &        & 0.72 & 91.86 \\
\cmark & \cmark &        & 0.84 & 92.61 \\
\cmark &        & \cmark & 0.84 & 92.38 \\
\rowcolor{lightgreen}\cmark & \cmark & \cmark & 0.84 & \textbf{93.48} \\
\bottomrule
\end{tabular}%
}%
}
\end{minipage}
\hfill
\begin{minipage}[t]{0.28\textwidth}
\centering
\vspace{-6pt}
\subcaption{The effect of inserted layers.}
\label{tab:ablation_layers}
\resizebox{\linewidth}{!}{%
\renewcommand{\arraystretch}{1.01}%
\begin{tabular}{lcc}
\toprule
Layers & \#TP (M) & PB\_T50\_RS \\
\midrule
    & 0.27     & 80.84      \\
\midrule
1→3    & 0.50     & 88.74      \\
1→6    & 0.61     & 90.12      \\
1→9    & 0.72     & 91.56      \\
\rowcolor{lightgreen}1→12   & 0.84     & \textbf{93.48}      \\
4→6    & 0.50     & 89.21      \\
4→9    & 0.61     & 90.37      \\
4→12   & 0.72     & 92.04      \\
7→9   & 0.50     & 88.63      \\
7→12   & 0.61     & 89.72      \\
10→12   & 0.50     & 88.11      \\
\bottomrule
\end{tabular}%
}
\end{minipage}
\hfill
\begin{minipage}[t]{0.365\textwidth}
\centering
\vspace{-6pt}
\subcaption{The effect of different PEFT methods.}
\label{tab:ablation_components_right}
\resizebox{\linewidth}{!}{%
\setlength{\tabcolsep}{1pt}
\renewcommand{\arraystretch}{1.15}
\begin{tabular}{lcc}
\toprule
Methods  &\#TP (M) & PB\_T50\_RS \\
\midrule
Mamba3D &16.90 & 92.05 \\
\midrule
 + Adapter\citep{houlsby2019parameter}& 0.87 & 84.45 ({\color{down}- 7.60}) \\
 + Prefix tuning\citep{li2021prefix}&0.71 & 83.26 ({\color{down}- 8.79}) \\
 + LoRA\citep{hu2022lora} &0.91 & 83.19 ({\color{down}- 8.86}) \\
\midrule
 + VPT-Deep\citep{jia2022visual} &0.41 & 83.37 ({\color{down}- 8.68}) \\
 + AdaptFormer\citep{chen2022adaptformer} &0.87 & 84.13 ({\color{down}- 7.92}) \\
 + BI-AdaptFormer\citep{jie2023revisiting} &0.87 & 82.88 ({\color{down}- 9.17}) \\
 + ProDiaL\citep{ham2025parameter} &0.97 & 83.81 ({\color{down}- 8.24}) \\
 \midrule
 + PointLoRA\citep{wang2025pointlora} &0.77 & 85.36 ({\color{down}- 6.69}) \\
 + PointGST\citep{liang2024pointgst} &0.62 & 87.38 ({\color{down}- 4.67})\\
 \rowcolor{lightgreen}\textbf{+ Mantis} &0.84 & \textbf{93.48} ({\color{up}+ 1.43}) \\
\bottomrule
\end{tabular}%
}
\end{minipage}
\vspace{-12pt}
\end{table}

\vspace{-6pt}
\subsection{Ablation study}
\label{subsec:ablation}

We conduct ablation studies on the most challenging PB\_T50\_RS variant based on Mamba3D to investigate the rationale and effectiveness of Mantis. More results are provided in Appendix~\ref{app:add_ablation}.

\textbf{Ablation on different components.} We first study the contribution of each component in Mantis. Table~\ref{tab:ablation_components} shows that SAA is the primary source of performance improvement. DSCD alone provides modest improvement, whereas introducing SAA leads to substantial gains. Building on SAA, both feature consistency and prediction consistency bring further gains, and their combination achieves the best result. Overall, DSCD functions as a regularizer that complements SAA, while feature- and prediction-level consistency provide complementary benefits.

\textbf{Ablation on different inserted layers.} One straightforward way to further reduce the number of tunable parameters is to insert SAA into only a subset of encoder layers rather than the full network. However, as illustrated in Table~\ref{tab:ablation_layers}, restricting SAA to partial layers consistently degrades performance, regardless of whether the selected layers are shallow, intermediate, or deep. This observation suggests that effective adaptation of a frozen Mamba backbone requires state modulation across all encoder layers, rather than within a few isolated layers. When SAA is limited to middle or late layers, the performance drops noticeably, indicating that suboptimal state evolution formed in earlier stages cannot be adequately compensated by later-layer correction.

\begin{wrapfigure}{r}{0.4\textwidth}
  \vspace{-16pt}
\centering
\includegraphics[width=1\linewidth,height=0.8\linewidth]{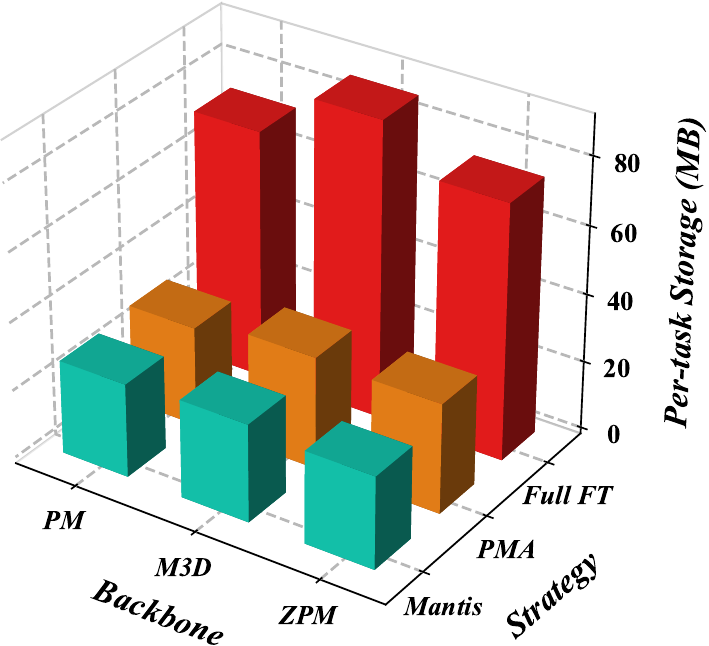} 
  \vspace{-16pt}
\caption{Per-task storage comparison.}
   \vspace{-16pt}
\label{fig:storage} 
\end{wrapfigure}

\textbf{Ablation on different PEFT methods.} To further validate the effectiveness of Mantis, we compare Mantis with several representative PEFT approaches originally developed for NLP\citep{houlsby2019parameter,li2021prefix,hu2022lora}, 2D vision\citep{jia2022visual,chen2022adaptformer,jie2023revisiting,ham2025parameter}, and 3D point clouds\citep{liang2024pointgst,wang2025pointlora}. As illustrated in Table~\ref{tab:ablation_components_right}, although these methods introduce different forms of parameter-efficient adaptation, they all suffer from substantial performance degradation compared with full fine-tuning. This suggests that existing PEFT strategies, even those tailored for point clouds, remain insufficient to effectively adapt frozen Mamba backbones. A possible reason is that most of them are designed for token-wise interactions or feature transformation, and therefore cannot properly align with the state-space dynamics and serialization-sensitive modeling mechanism of Mamba. In contrast, the strength of Mantis lies not merely in parameter efficiency, but in its Mamba-native design that better matches the intrinsic adaptation demands of selective SSMs.

\textbf{Analysis on storage efficiency.}
Since task-specific parameters must be independently maintained, per-task storage constitutes a critical scalability factor when adapting a shared pretrained backbone to multiple downstream tasks. As shown in Figure~\ref{fig:storage}, Mantis consistently incurs lower storage overhead across all three Mamba backbones. Full fine-tuning requires \emph{2.66$\times$, 3.03$\times$, and 2.76$\times$} the per-task storage of Mantis on PointMamba, Mamba3D, and ZigzagPointMamba, respectively. Moreover, Mantis further reduces storage by 5.8\%, 12.7\%, and 15.8\% compared with PMA\citep{zha2025pma}, demonstrating more favorable storage scalability for parameter-efficient downstream adaptation.

\section{Conclusion}

In this paper, we propose Mantis, the first Mamba-native PEFT framework for 3D PFMs. We identify the mismatch between existing Transformer-native PEFT methods and Mamba-based backbones, and address it through state-level adaptation and cross-serialization regularization. Extensive experiments demonstrate that Mantis achieves strong downstream performance while training less than 5\% of the backbone parameters. As an early exploration of Mamba-native PEFT, we expect Mantis can serve as a baseline and provide insights for future 3D PEFT research. Building on this foundation, extending Mamba-native PEFT to complementary modalities remains a promising but underexplored direction.

\section{Statements}

\subsection*{AI Use Statement}
Generative AI tools were used only to assist with language editing and presentation refinement.
All technical content, experimental results, analyses, and conclusions were produced and verified by the authors, who take full responsibility for the final manuscript.

\subsection*{Ethics Statement}
This work is based on publicly available 3D point cloud benchmarks and does not involve human-subject studies or the collection of private or personally identifiable information.
We are not aware of any additional ethical concerns arising from this research.

\subsection*{Reproducibility Statement}
We provide detailed descriptions of the model architecture, training settings, evaluation protocols, and implementation details in the main paper and appendix.
Experiments follow the standard dataset splits and evaluation settings of the corresponding benchmarks, and our reported results are averaged over three random seeds where applicable.
The source code and configuration files are publicly available to ensure the reproducibility of our results.

\bibliography{iclr2027_conference}
\bibliographystyle{iclr2027_conference}

\clearpage

\appendix

\startcontents[appendix]

\section*{Appendix Table of Contents}
\begingroup
\hypersetup{
  linkcolor=black,
  pdfborder={0 0 0}
}
\setcounter{tocdepth}{2}

\titlecontents{section}
  [2.7em]
  {\vspace{1.2em}\bfseries}
  {\contentslabel{2.2em}}
  {}
  {\titlerule*[0.6em]{.}\contentspage}

\titlecontents{subsection}
  [3.5em]
  {\vspace{0.4em}\small}
  {\contentslabel{3.0em}}
  {}
  {\titlerule*[0.6em]{.}\contentspage}

\printcontents[appendix]{}{1}{}
\endgroup

\clearpage

\section{Additional theoretical analysis}
\label{app:theory}

\subsection{Preliminaries: State Space Models and PointMamba}
\label{app:preliminaries}

Drawing inspiration from control theory, State Space Models (SSMs) map an input sequence $x_t$ to an output $y_t$ through a latent state $h_t$. In its discretized form, the transformation can be written as:
\begin{equation}
h_t = \overline{\boldsymbol{A}} h_{t-1} + \overline{\boldsymbol{B}} x_t, \qquad 
y_t = \boldsymbol{C} h_t,
\label{eq:ssm}
\end{equation}
where $\overline{\boldsymbol{A}}$ and $\overline{\boldsymbol{B}}$ are the discretized state transition and input projection matrices. While early structured SSMs rely on fixed transition parameters, Mamba (S6)\citep{gu2023mamba} introduces input-dependent parameters $(\boldsymbol{B}, \boldsymbol{C}, \boldsymbol{\Delta})$, enabling selective and content-aware sequence modeling.

For point cloud analysis, PointMamba\citep{liang2024pointmamba} adopts the SSM framework by treating serialized point tokens as sequential inputs. The state update can be written as:
\begin{equation}
h_t = \overline{\boldsymbol{A}}_t h_{t-1} + \overline{\boldsymbol{B}}_t x_t, \qquad 
y_t = \boldsymbol{C}_t h_t,
\end{equation}
where $h_t$ denotes the hidden state and $x_t$ denotes the input point token. Mamba keeps the structured continuous-time transition $\boldsymbol{A}$ fixed, while $\boldsymbol{B}_t$, $\boldsymbol{C}_t$, and $\boldsymbol{\Delta}_t$ are input-dependent. Accordingly, the discretized transition $\overline{\boldsymbol{A}}_t=\exp(\boldsymbol{A}\boldsymbol{\Delta}_t)$ remains input-dependent through $\boldsymbol{\Delta}_t$. This selective state update provides efficient sequence modeling, but also causes the hidden-state propagation to depend on the serialized point order.

\subsection{A closer look at selective SSM under serialization}
\label{app:selective_ss}

Different from LTI SSMs with fixed transition kernels, Mamba keeps the
structured continuous-time transition $\boldsymbol{A}$ fixed while using
input-dependent selective coefficients $\boldsymbol{B}_t$,
$\boldsymbol{C}_t$, and $\boldsymbol{\Delta}_t$. Consequently, the
effective discretized transition
$\overline{\boldsymbol{A}}_t
=\exp(\boldsymbol{A}\boldsymbol{\Delta}_t)$
also becomes input-dependent through $\boldsymbol{\Delta}_t$.
For point cloud modeling, different valid serializations therefore induce
different input sequences and selective operator chains, making the
effective transfer kernel dependent on the chosen serialization.

As described in Sec.~\ref{app:preliminaries}, for branch $k \in \{1,2\}$, the selective state-space model updates the hidden state according to
\begin{equation}
h_t^{(k,0)}
=
\overline{\boldsymbol{A}}_t^{(k)} h_{t-1}^{(k,0)}
+
\overline{\boldsymbol{B}}_t^{(k)} x_t^{(k)},
\qquad
y_t^{(k,0)}
=
\boldsymbol{C}_t^{(k)} h_t^{(k,0)},
\label{eq:app_ssm_recursion}
\end{equation}

where $x_t^{(k)} \in \mathbb{R}^{d}$ denotes the serialized input token,
$\overline{\boldsymbol{A}}_t^{(k)}$ and
$\overline{\boldsymbol{B}}_t^{(k)}$ are the discretized operators of the
frozen selective SSM, and $\boldsymbol{C}_t^{(k)}$ is the corresponding
input-dependent readout coefficient.

Ignoring the initial state for simplicity, the recurrent form in Eq.~\eqref{eq:app_ssm_recursion} can be expanded as:
\begin{equation}
h_t^{(k,0)}
=
\sum_{i=1}^{t}
\left(
\prod_{j=i+1}^{t}\overline{\boldsymbol{A}}_{j}^{(k)}
\right)
\overline{\boldsymbol{B}}_{i}^{(k)} x_i^{(k)},
\label{eq:app_hidden_expand}
\end{equation}
where the product is defined in temporal order, \emph{i.e.},
\(
\prod_{j=i+1}^{t}\overline{\boldsymbol{A}}_{j}^{(k)}
=
\overline{\boldsymbol{A}}_{t}^{(k)} \overline{\boldsymbol{A}}_{t-1}^{(k)} \cdots \overline{\boldsymbol{A}}_{i+1}^{(k)}
\),
and equals the identity matrix when $i=t$.

Substituting Eq.~\eqref{eq:app_hidden_expand} into the readout equation yields:
\begin{equation}
y_t^{(k,0)}
=
\sum_{i=1}^{t}
\boldsymbol{C}_{t}^{(k)}
\left(
\prod_{j=i+1}^{t}\overline{\boldsymbol{A}}_{j}^{(k)}
\right)
\overline{\boldsymbol{B}}_{i}^{(k)} x_i^{(k)}.
\label{eq:app_output_expand}
\end{equation}

Thus, the selective transfer kernel from step $i$ to step $t$ is defined as:
\begin{equation}
\boldsymbol{W}_{t,i}^{(k,0)}
=
\boldsymbol{C}_{t}^{(k)}
\left(
\prod_{j=i+1}^{t}\overline{\boldsymbol{A}}_{j}^{(k)}
\right)
\overline{\boldsymbol{B}}_{i}^{(k)},
\qquad 1 \le i \le t \le n,
\label{eq:app_transfer_kernel_frozen}
\end{equation}
where $n$ is the serialized sequence length. Then Eq.~\eqref{eq:app_output_expand} can be rewritten as:
\begin{equation}
y_t^{(k,0)}
=
\sum_{i=1}^{t}
\boldsymbol{W}_{t,i}^{(k,0)} x_i^{(k)}.
\label{eq:app_output_transfer}
\end{equation}

Stacking all outputs $\hat{\boldsymbol y}^{(k,0)} = [y_1^{(k,0)},\dots,y_n^{(k,0)}]^\top$ and inputs $\hat{\boldsymbol x}^{(k)} = [x_1^{(k)},\dots,x_n^{(k)}]^\top$, Eq.~\eqref{eq:app_output_transfer} can be expressed in matrix form:
\begin{equation}
\hat{\boldsymbol y}^{(k,0)}
=
\boldsymbol{W}^{(k,0)} \hat{\boldsymbol x}^{(k)},
\qquad
\boldsymbol{W}^{(k,0)}
=
\begin{bmatrix}
\boldsymbol{W}_{1,1}^{(k,0)} & 0 & \cdots & 0 \\
\boldsymbol{W}_{2,1}^{(k,0)} & \boldsymbol{W}_{2,2}^{(k,0)} & \cdots & 0 \\
\vdots & \vdots & \ddots & \vdots \\
\boldsymbol{W}_{n,1}^{(k,0)} & \boldsymbol{W}_{n,2}^{(k,0)} & \cdots & \boldsymbol{W}_{n,n}^{(k,0)}
\end{bmatrix}.
\label{eq:app_transfer_matrix}
\end{equation}
Since $\boldsymbol{W}^{(k,0)}$ is lower triangular, the selective SSM defines a causal, input-dependent transfer matrix over the serialized point sequence.

Eq.~\eqref{eq:app_transfer_matrix} also reveals the source of serialization sensitivity. For two valid serializations of the same point cloud, not only the input order $\hat{\boldsymbol x}^{(1)}$ and $\hat{\boldsymbol x}^{(2)}$ differ, but the induced operator chains
\(
\prod_{j=i+1}^{t}\overline{\boldsymbol{A}}_{j}^{(1)}
\)
and
\(
\prod_{j=i+1}^{t}\overline{\boldsymbol{A}}_{j}^{(2)}
\)
also differ. Consequently, the same underlying geometry may lead to different transfer matrices and hence different representations under distinct valid serializations.

\subsection{Effect of State-Aware Adapter on selective state-space dynamics}
\label{app:saa_kernel}

To characterize the state-level adaptation mechanism of SAA, we analyze its effect on the selective transfer kernel of the frozen Mamba backbone.
Building on the selective transfer kernel derived in Appendix~\ref{app:selective_ss}, $\boldsymbol{W}_{t,i}^{(k)}$ denotes the selective transfer kernel after SAA modulation, while $\boldsymbol{W}_{t,i}^{(k,0)}$ denotes its frozen counterpart without SAA:
\begin{equation}
\boldsymbol{W}_{t,i}^{(k,0)}
=
\boldsymbol{C}_{t}^{(k)}
\left(
\prod_{j=i+1}^{t}\overline{\boldsymbol{A}}_{j}^{(k)}
\right)
\overline{\boldsymbol{B}}_{i}^{(k)}.
\label{eq:app_transfer_kernel}
\end{equation}
where $\overline{\boldsymbol{A}}_{t}^{(k)}$ and $\overline{\boldsymbol{B}}_{t}^{(k)}$ are the discretized operators of the original frozen selective SSM, and $\boldsymbol{C}_{t}^{(k)}$ is the corresponding readout matrix.
Under SAA, we first define the coefficient-specific factorized correction as:
\begin{equation}
\boldsymbol{\Gamma}_{O,t}^{(k)}
=
\mathbf{U}_{O}
\operatorname{diag}\!\bigl(\mathbf{u}_t^{(k)}\bigr)
\mathbf{V}_{O},
\qquad
O\in\{B,C,\Delta\},
\label{eq:app_factorized_correction}
\end{equation}
where $\mathbf{u}_t^{(k)} \in \mathbb{R}^{r}$ is the shared sparse
control signal, and $\mathbf{U}_{O}$ and $\mathbf{V}_{O}$ are
coefficient-specific low-rank modulation matrices. The actual
native-layout correction is then given by
\begin{equation}
\delta_{O,t}^{(k)}
=
\mathcal{R}_{O}\!\left(
\boldsymbol{\Gamma}_{O,t}^{(k)}
\right),
\qquad
\widetilde{\boldsymbol{O}}_t^{(k)}
=
\boldsymbol{O}_t^{(k)}
+
\delta_{O,t}^{(k)}.
\label{eq:app_native_correction}
\end{equation}
Here, $\mathcal{R}_{O}(\cdot)$ maps the factorized correction to the
native coefficient layout. The structured continuous-time transition
$\boldsymbol{A}$ remains frozen.
For completeness, in the standard Mamba selective scan,
$\boldsymbol{B}_t$ and $\boldsymbol{C}_t$ are token-wise state
coefficients in $\mathbb{R}^{d_s}$, while
$\boldsymbol{\Delta}_t\in\mathbb{R}^{d_{\mathrm{in}}}$ is defined over
the inner channels, where $d_s$ and $d_{\mathrm{in}}$ denote the SSM
state size and Mamba inner dimension, respectively. Accordingly,
$\mathcal{R}_{O}(\cdot)$ is implemented as a parameter-free
reshape/permutation that converts the factorized correction to the native
tensor layout of coefficient $O$ before residual addition.
\begin{equation}
\boldsymbol{W}_{t,i}^{(k)}
=
\widetilde{\boldsymbol{C}}_{t}^{(k)}
\left(
\prod_{j=i+1}^{t}\widehat{\boldsymbol{A}}_{j}^{(k)}
\right)
\widehat{\boldsymbol{B}}_{i}^{(k)}.
\label{eq:app_controlled_kernel}
\end{equation}
The corresponding kernel perturbation is
\begin{equation}
\Delta \boldsymbol{W}_{t,i}^{(k)}
=
\boldsymbol{W}_{t,i}^{(k)} - \boldsymbol{W}_{t,i}^{(k,0)}.
\label{eq:app_kernel_difference}
\end{equation}
For each selective coefficient $O$, branch $k$, and sequence position $t$,
the factorized correction $\boldsymbol{\Gamma}_{O,t}^{(k)}$ is a sum of
at most $|\mathbf{u}_t^{(k)}|_0$ active rank-one components. Therefore,
\begin{equation}
\operatorname{rank}\!\left(
\boldsymbol{\Gamma}_{O,t}^{(k)}
\right)
\le
|\mathbf{u}_t^{(k)}|_0
\le r.
\label{eq:app_rank_bound}
\end{equation}
Therefore, SAA constructs each native coefficient correction from a sparse
low-rank factorized source. The low-rank property applies to
$\boldsymbol{\Gamma}_{O,t}^{(k)}$, while
$\mathcal{R}_{O}(\cdot)$ only maps this structured correction to the
native coefficient layout. Through discretization and recurrent
propagation, these native-layout corrections induce structured changes
in the selective transfer kernel and pretrained state dynamics. This
factorized parameterization restricts trainable adaptation to a
low-dimensional subspace and explains the parameter efficiency of SAA.

We further examine how the resulting native-layout coefficient corrections
affect the hidden-state trajectory through discretization and recurrence.
The hidden states produced by the frozen backbone and the SAA-modulated
backbone are denoted as $h_t^{(k,0)}$ and $h_t^{(k)}$, respectively:
\begin{equation}
h_t^{(k,0)}
=
\overline{\boldsymbol{A}}_t^{(k)} h_{t-1}^{(k,0)}
+
\overline{\boldsymbol{B}}_t^{(k)} x_t^{(k)},
\qquad
h_t^{(k)}
=
\widehat{\boldsymbol{A}}_t^{(k)} h_{t-1}^{(k)}
+
\widehat{\boldsymbol{B}}_t^{(k)} x_t^{(k)}.
\end{equation}
The hidden-state deviation
$\delta h_t^{(k)}=h_t^{(k)}-h_t^{(k,0)}$ satisfies
\begin{equation}
\begin{aligned}
\delta h_t^{(k)}
&=
\overline{\boldsymbol{A}}_t^{(k)}\delta h_{t-1}^{(k)}
+
\left(
\widehat{\boldsymbol{A}}_t^{(k)}
-
\overline{\boldsymbol{A}}_t^{(k)}
\right)h_{t-1}^{(k)}
\\
&\quad+
\left(
\widehat{\boldsymbol{B}}_t^{(k)}
-
\overline{\boldsymbol{B}}_t^{(k)}
\right)x_t^{(k)} .
\end{aligned}
\end{equation}
Thus, the recurrent deviation is propagated by the frozen transition
$\overline{\boldsymbol{A}}_t^{(k)}$, while SAA contributes an additive
transition perturbation. Assume that
$\|\overline{\boldsymbol{A}}_t^{(k)}\|_2\le\rho<1$,
$\|h_t^{(k)}\|_2\le H$, and $\|x_t^{(k)}\|_2\le X$.
Since $\boldsymbol{A}$ remains frozen and
$\widehat{\boldsymbol{A}}_t^{(k)}
=\exp(\boldsymbol{A}\widetilde{\boldsymbol{\Delta}}_t^{(k)})$,
a bounded correction
$\|\widetilde{\boldsymbol{\Delta}}_t^{(k)}
-\boldsymbol{\Delta}_t^{(k)}\|_2\le\varepsilon_{\Delta}$
induces, over a bounded discretization domain,
\begin{equation}
\left\|
\widehat{\boldsymbol{A}}_t^{(k)}
-
\overline{\boldsymbol{A}}_t^{(k)}
\right\|_2
\le
\varepsilon_{\mathrm{tr}},
\qquad
\left\|
\widehat{\boldsymbol{B}}_t^{(k)}
-
\overline{\boldsymbol{B}}_t^{(k)}
\right\|_2
\le
\varepsilon_B ,
\end{equation}
where $\varepsilon_{\mathrm{tr}}=O(\varepsilon_{\Delta})$ follows from
the local Lipschitz continuity of the matrix exponential.

The hidden-state deviation is therefore bounded by
\begin{equation}
\left\|
\delta h_t^{(k)}
\right\|_2
\le
\rho
\left\|
\delta h_{t-1}^{(k)}
\right\|_2
+
\varepsilon_{\mathrm{tr}}H
+
\varepsilon_BX ,
\label{eq:state_deviation_recursion}
\end{equation}
which yields
\begin{equation}
\left\|
\delta h_t^{(k)}
\right\|_2
\le
\rho^t
\left\|
\delta h_0^{(k)}
\right\|_2
+
\frac{1-\rho^t}{1-\rho}
\left(
\varepsilon_{\mathrm{tr}}H
+
\varepsilon_BX
\right).
\label{eq:bounded_hidden_deviation}
\end{equation}

The modulation of $\boldsymbol{C}_t$ affects the output rather than the
hidden-state recurrence. If
$\|\boldsymbol{C}_t^{(k)}\|_2\le M_C$ and
$\|\widetilde{\boldsymbol{C}}_t^{(k)}
-\boldsymbol{C}_t^{(k)}\|_2\le\varepsilon_C$, then
\begin{equation}
\left\|
y_t^{(k)}-y_t^{(k,0)}
\right\|_2
\le
M_C\left\|\delta h_t^{(k)}\right\|_2
+
\varepsilon_C H .
\label{eq:bounded_output_deviation}
\end{equation}
Therefore, under the stated boundedness assumptions, SAA induces bounded
perturbations in both the hidden-state trajectory and the selective-SSM
output. This provides a conditional perturbation analysis of the adapted
state dynamics, while the optimization stability observed in practice is
supported separately by the empirical training dynamics.

\subsection{Parameter and complexity analysis}
\label{app:complexity}

We finally analyze the additional parameter and computational overhead introduced by the proposed State-Aware Adapter (SAA). 
Recall that SAA consists of the pre-scan control projections
$W_x^{\phi}$, $W_z^{\phi}$, and $W_e^{\phi}$, the control-feature
fusion process $\Phi(\cdot)$, the driving and gating projections
$W_{\mathrm{drv}}^{\psi}$ and $W_{\mathrm{gt}}^{\psi}$, and the
coefficient-specific low-rank modulation matrices $\mathbf{U}_{O}$ and
$\mathbf{V}_{O}$. Accordingly, the total number of trainable parameters in one SAA module is
\begin{equation}
|\Theta_{\mathrm{SAA}}|
=
|W_x^{\phi}|
+
|W_z^{\phi}|
+
|W_e^{\phi}|
+
|\Theta_{\Phi}|
+
|W_{\mathrm{drv}}^{\psi}|
+
|W_{\mathrm{gt}}^{\psi}|
+
\sum_{O\in\{B,C,\Delta\}}
\left(
|\mathbf{U}_{O}|+|\mathbf{V}_{O}|
\right).
\label{eq:app_param_count}
\end{equation}
Specifically,
\(
W_x^{\phi}, W_z^{\phi}, W_e^{\phi}
\in\mathbb{R}^{d_{\phi}\times d}.
\)
and
\(
W_{\mathrm{drv}}^{\psi}, W_{\mathrm{gt}}^{\psi}
\in\mathbb{R}^{r\times d_{\phi}}.
\)
In our implementation, the control-feature fusion process $\Phi(\cdot)$
is instantiated as a lightweight MLP that projects the concatenated $3d_{\phi}$-dimensional control feature into the shared $d_{\phi}$-dimensional control space. For each selective coefficient $O\in\{B,C,\Delta\}$, we denote
\(
\mathbf{U}_{O}\in\mathbb{R}^{d_O^{\mathrm{out}}\times r}
\)
and
\(
\mathbf{V}_{O}\in\mathbb{R}^{r\times d_O^{\mathrm{in}}},
\)
where $d_O^{\mathrm{out}}$ and $d_O^{\mathrm{in}}$ follow the native
dimensions of selective coefficient $O$, and $r$ is the bottleneck dimension. The modulation preserves the native layout of each selective coefficient
without introducing dense transition matrices. The total number of trainable parameters in one SAA module therefore becomes
\begin{equation}
|\Theta_{\mathrm{SAA}}|
=
3d_{\phi}d
+
3d_{\phi}^{2}+d_{\phi}
+
2rd_{\phi}
+
\sum_{O\in\{B,C,\Delta\}}
r\left(
d_O^{\mathrm{out}}+d_O^{\mathrm{in}}
\right).
\label{eq:app_param_final}
\end{equation}
Eq.~\eqref{eq:app_param_final} shows that the adaptation overhead is dominated by low-dimensional control projections and low-rank modulation. Since $r \ll d$ and $d_{\phi}$ is small, the number of trainable parameters introduced by SAA remains much smaller than that of full fine-tuning.

We next analyze the computational overhead. SAA introduces only
token-wise control-feature projection, control-feature fusion, sparse
control generation, and factorized coefficient modulation. All these
operations are computed independently across sequence positions, so the
additional overhead scales linearly with the sequence length $n$.
More importantly, all SAA coefficients are generated in parallel before
recurrence and consumed by a single native selective-scan call per
SAA-Mamba block and serialization branch. Hence, SAA introduces no
additional recurrent pass and preserves the native linear-time
selective-scan structure.
\section{More experimental results}
\label{sec:app_more}
\subsection{Training detail}
\label{app:train_detail}
\renewcommand{\arraystretch}{1.1}  % 放在表格前或环境内
\begin{table*}[htbp]
  \vspace{-8pt}
\centering
\caption{Implementation details for various downstream tasks.}
\vspace{-8pt}
\label{tab:implementation}

\small
\setlength{\tabcolsep}{5pt}
\renewcommand{\arraystretch}{1.15}

\resizebox{\textwidth}{!}{%
\begin{tabular}{lccccc}
\toprule

\multirow{2}{*}{Configuration}
& \multicolumn{2}{c}{Classification}
& \multicolumn{3}{c}{Segmentation}
\\

\cmidrule(lr){2-3}
\cmidrule(lr){4-6}

& ModelNet40
& ScanObjectNN
& ShapeNetPart
& S3DIS Area5
& ScanNet
\\

\midrule

Optimizer
& AdamW
& AdamW
& AdamW
& AdamW
& AdamW
\\

Learning rate
& 3e-4
& 5e-4
& 2e-4
& 1e-4
& 1e-3
\\

Weight decay
& 5e-2
& 5e-2
& 5e-2
& 5e-2
& 5e-2
\\

Learning rate scheduler
& cosine
& cosine
& cosine
& cosine
& cosine
\\

Training epochs
& 300
& 300
& 300
& 100
& 200
\\

Warm-up epochs
& 10
& 10
& 10
& 5
& 10
\\

Batch size
& 32
& 32
& 16
& 4
& 4
\\

$r$ in State-Aware Adapter
& 8
& 8
& 8
& 8
& 8
\\

Softmax temperature $\tau$
& 1
& 1
& 1
& 1
& 1
\\

Num. of encoder layers
& 12
& 12
& 12
& 12
& 12
\\

Input points
& 1024
& 2048
& 2048
& 24576
& 32768
\\

Num. of patches
& 64
& 128
& 128
& 192
& 256
\\

Patch size
& 32
& 32
& 32
& 48
& 48
\\

Augmentation
& Scale\&Trans/Rotation
& Scale\&Trans
& Scale\&Trans
& Scale\&Trans/Rotation
& Scale\&Trans/Rotation
\\

\bottomrule
\end{tabular}%
}
\vspace{-4pt}
\end{table*}

All experiments, including classification, few-shot, and segmentation, follow the standard dataset splits as defined in the respective benchmarks, and results are averaged over multiple runs with different random seeds to ensure reproducibility. For comparison, full fine-tuning retains the native backbone architecture, whereas the lightweight order-conditioning and dual-serialization components are specific to Mantis and are fully included in its reported parameter and computational budgets. 

We adopt downstream fine-tuning configurations in alignment with the pioneering work PointMamba\citep{liang2024pointmamba}, as in Table~\ref{tab:implementation}. For example, when fine-tuning on ModelNet40\citep{wu20153d}, the training process spans 300 epochs, using a cosine learning rate scheduler that starts at 3e-4, with a 10-epoch warm-up period. The AdamW optimizer is employed. All experiments are conducted on a single GeForce RTX 5090 using PyTorch version 2.7.1+cu128.

During training, each point cloud is instantiated with two valid serialization orders.
The two branches share the backbone and SAA parameters and are jointly used to compute the task objective and the feature- and prediction-level consistency terms in DSCD.
At inference, we retain only a single serialization branch and discard the auxiliary serialization used for consistency regularization.
Therefore, Mantis does not require dual-pass inference, prediction ensembling, or test-time voting.
All reported inference FLOPs and latency are measured under this single-serialization protocol and include the additional computation introduced by SAA.
In contrast, the reported training time and peak memory account for the complete dual-serialization training procedure with DSCD.

\begin{table*}[t]
\centering
\small
\caption{Ablation study on Mantis based on Mamba3D\citep{li2024mamba3d}, including sparse controller design, bottleneck dimension $r$, and scanning curves.}
\label{tab:extra_ablation}
\begin{minipage}[t]{0.39\textwidth}
\centering
  \vspace{-8pt}
\subcaption{The effect of different scanning curves.}
\label{tab:scan_curve_pb}
\renewcommand{\arraystretch}{1}
\resizebox{\linewidth}{!}{%
\begin{tabular}{lc}
\toprule
Scanning curve & PB\_T50\_RS (\%) \\
\midrule
Random & 91.74 \\
\rowcolor{lightgreen} Hilbert and Trans-Hilbert & \textbf{93.48} \\
Z-order and Trans-Z-order & 92.86 \\
Hilbert and Z-order & 93.02 \\
Trans-Hilbert and Trans-Z-order & \underline{93.21} \\
\bottomrule
\end{tabular}%
}
\end{minipage}
\hfill
\begin{minipage}[t]{0.25\textwidth}
\centering
  \vspace{-8pt}
\subcaption{The effect of $r$.}
\label{tab:r_trend}
\renewcommand{\arraystretch}{1.08}
\resizebox{\linewidth}{!}{%
\begin{tabular}{ccc}
\toprule
$r$ & \#TP (M) & PB\_T50\_RS (\%) \\
\midrule
\rowcolor{lightgreen}8  & 0.84 & 93.48 \\
16 & 1.00 & \underline{93.50} \\
32 & 1.32 & \textbf{93.58} \\
64 & 1.95 & 93.47 \\
72 & 2.12 & 93.41 \\
\bottomrule
\end{tabular}%
}
\end{minipage}
\hfill
\begin{minipage}[t]{0.32\textwidth}
\centering
  \vspace{-8pt}
\subcaption{The effect of controller design.}
\label{tab:controller_design}
\renewcommand{\arraystretch}{1.13}
\resizebox{\linewidth}{!}{%
\begin{tabular}{lcc}
\toprule
Controller & \#TP (M) & PB\_T50\_RS (\%) \\
\midrule
Dense-MLP & 0.95 & 92.71 \\
Sigmoid Gate & 0.84 & 92.88 \\
Tanh Gate & 0.84 & 92.97 \\
Hard-threshold & 0.84 & \underline{93.16} \\
\rowcolor{lightgreen} Soft-threshold & 0.84 & \textbf{93.48} \\
\bottomrule
\end{tabular}%
}
\end{minipage}
  \vspace{-8pt}
\end{table*}

\begin{table*}[t]
\centering
\small
\caption{Ablation study on the internal design of SAA based on Mamba3D\citep{li2024mamba3d}, including state fusion function and modulated state-space operators.}
\label{tab:saa_additional_ablation}
\begin{minipage}[t]{0.50\textwidth}
\centering
  \vspace{-8pt}
\subcaption{Ablation on fusion function $\Phi(\cdot)$.}
\label{tab:fusion_function}
\renewcommand{\arraystretch}{1.18}
\resizebox{\linewidth}{!}{%
\begin{tabular}{lcc}
\toprule
Fusion function $\Phi(\cdot)$ & \#TP (M) & PB\_T50\_RS (\%) \\
\midrule
Addition & 0.81 & 92.34 \\
Concatenation & 0.82 & 92.71 \\
Gated Fusion & 0.86 & \underline{93.21} \\
Cross-attention Fusion & 0.91 & 93.12 \\
\rowcolor{lightgreen} Concat. + MLP & 0.84 & \textbf{93.48} \\
\bottomrule
\end{tabular}%
}
\end{minipage}
\hfill
\begin{minipage}[t]{0.46\textwidth}
\centering
  \vspace{-8pt}
\subcaption{Ablation on modulated operators.}
\label{tab:operator_target}
\renewcommand{\arraystretch}{1.08}
\resizebox{\linewidth}{!}{%
\begin{tabular}{lcc}
\toprule
Modulated operators & \#TP (M) & PB\_T50\_RS (\%) \\
\midrule
$\boldsymbol{B}_t$ only & 0.74 & 92.08 \\
$\boldsymbol{C}_t$ only & 0.74 & 92.43 \\
$\boldsymbol{\Delta}_t$ only & 0.76 & 92.31 \\
$\boldsymbol{B}_t, \boldsymbol{C}_t$ & 0.78 & 92.67 \\
$\boldsymbol{B}_t, \boldsymbol{\Delta}_t$ & 0.80 & \underline{92.96} \\
\rowcolor{lightgreen} $\boldsymbol{B}_t, \boldsymbol{C}_t, \boldsymbol{\Delta}_t$ & 0.84 & \textbf{93.48} \\
\bottomrule
\end{tabular}%
}
\end{minipage}
  \vspace{-8pt}
\end{table*}
  \vspace{-8pt}
\subsection{Additional ablation study}
\label{app:add_ablation}
\textbf{Ablation study on different scanning curves.} We further evaluate DSCD under different scanning curves, including Hilbert, Morton (Z-order), and their transposed variants. As shown in Table~\ref{tab:extra_ablation} (a), all structured scanning strategies consistently outperform random serialization on the hardest PB\_T50\_RS setting, indicating that geometry-aware traversal orders are important for stable state propagation in Mamba-based backbones. Moreover, since DSCD enforces consistency across two serialization branches, the scanning curves affect not only the sequence modeling quality within each branch, but also the complementarity between the two views. Among all candidates, using Hilbert and Trans-Hilbert as the two training views achieves the best performance, likely because it provides a favorable balance between locality preservation and cross-serialization diversity.

\begin{wrapfigure}{r}{0.36\textwidth}
\centering
 \vspace{-16pt}
\includegraphics[width=1\linewidth]{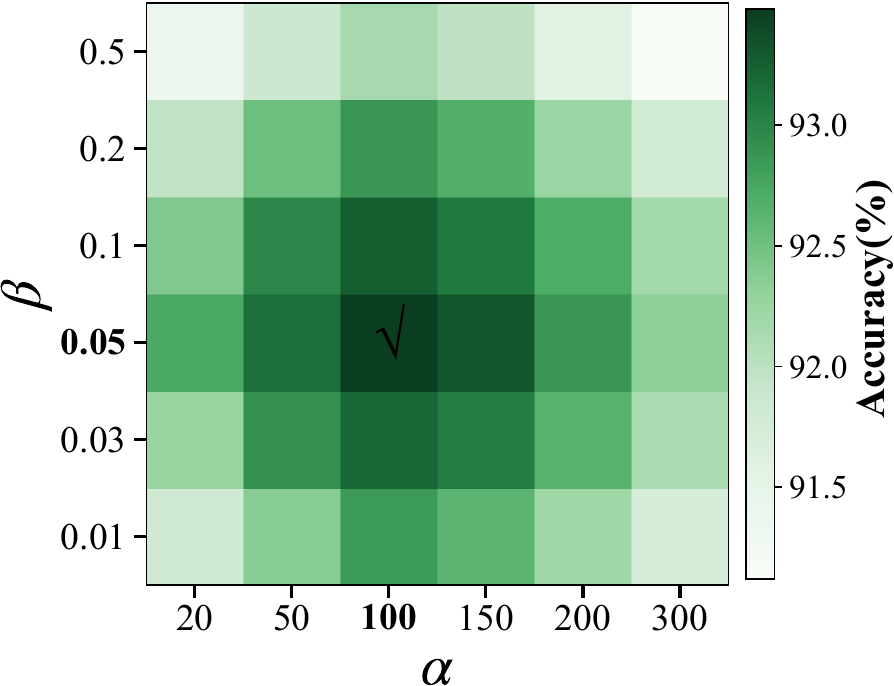} 
  \vspace{-20pt}
\caption{Sensitivity analysis of the DSCD hyperparameters $\alpha$ and $\beta$.}
  \vspace{-12pt}
\label{fig:heat} 
\end{wrapfigure}

\textbf{Ablation study on bottleneck dimension $r$.} The SAA aims to reduce the number of trainable parameters during fine-tuning, and the bottleneck dimension $r$ balances adaptation efficiency and downstream performance. As shown in Table~\ref{tab:extra_ablation} (b), while increasing $r$ moderately enlarges the adaptation capacity of SAA, the gain over $r=8$ remains marginal, suggesting that a small bottleneck is sufficient to capture most useful task-specific modulation. Therefore, we adopt $r=8$ as the default setting, since it provides the best trade-off between accuracy and parameter efficiency.

\textbf{Ablation study on controller design.} To examine whether the sparse proximal controller is necessary, we compare the proposed soft-thresholding controller with several dense or hard-selection alternatives under matched trainable parameters. As shown in Table~\ref{tab:extra_ablation} (c), replacing the soft-thresholding controller with Dense-MLP, Sigmoid Gate, Tanh Gate, or Hard-threshold consistently degrades performance. This indicates that the improvement of SAA lies not merely in low-rank dynamic parameterization, but also in sparse and magnitude-controlled perturbations induced by soft-thresholding. Therefore, the soft-thresholding controller provides a more stable way to modulate frozen Mamba backbones.

\textbf{Ablation study on fusion function $\Phi(\cdot)$.}
We study choices for the fusion function $\Phi(\cdot)$ in SAA. As shown in Table~\ref{tab:saa_additional_ablation} (a), simple addition or concatenation provides limited fusion capacity, while gated fusion and cross-attention achieve better results with more trainable parameters. In comparison, the proposed MLP-based fusion function achieves the best performance, suggesting that nonlinear fusion of the projected control features is beneficial for generating effective state-aware control signals.

\textbf{Ablation study on modulated operators.}
We also examine which state-space operators should be modulated by SAA. As shown in Table~\ref{tab:saa_additional_ablation} (b), modulating a single operator performs worse, indicating that adapting only one part of the selective SSM is insufficient. Jointly modulating $\boldsymbol{B}_t$, $\boldsymbol{C}_t$, and $\boldsymbol{\Delta}_t$ achieves the best performance, indicating that input injection, state readout, and adaptive discretization provide complementary adaptation capacity. We therefore retain the structured transition operator $\boldsymbol{A}$ frozen and apply SAA only to the native input-dependent selective coefficients.

\textbf{Controlled analysis of state-aware adaptation.}
To disentangle the effects of state-aware control, dual-serialization training, and consistency regularization, we conduct controlled comparisons in Table~\ref{tab:state_control_ablation}. Full fine-tuning benefits marginally from dual serialization and moderately from DSCD, suggesting that the improvement of Mantis is not primarily caused by the dual-view training protocol. Adapting the native projections generating $B$, $C$, and $\Delta$ substantially improves LoRA, showing that conventional PEFT can indeed affect selective state dynamics when appropriately inserted. However, this SSM-oriented LoRA remains below SAA, even when equipped with DSCD. Furthermore, removing the incoming block context $z_t$ from the pre-scan controller causes a noticeable degradation under the same parameter budget. Together, these results indicate that contextualized pre-scan conditioning strengthens SAA beyond input-only dynamic modulation, with DSCD providing regularization.

\textbf{Hyperparameter sensitivity analysis.}
We analyze the sensitivity of DSCD to the loss weights $\alpha$ and $\beta$.
Figure~\ref{fig:heat} shows that Mantis is relatively stable across a broad range of hyperparameters, indicating that the effectiveness of DSCD is not restricted to a single carefully tuned setting.
The best performance occurs around $\alpha=100$ and $\beta=0.05$, which are adopted as the default hyperparameters.

\begin{table*}[t]
\centering
\vspace{-4pt}

% =========================================================
% Left: Controlled state-aware adaptation
% =========================================================
\begin{minipage}[t]{0.51\textwidth}
\vspace{0pt}
\centering

\captionof{table}{
Controlled comparison of state-aware adaptation and dual-serialization regularization on Mamba3D.
LoRA ($B,C,\Delta$) adapts native SSM projections under a matched parameter budget.}
\label{tab:state_control_ablation}

\small
\renewcommand{\arraystretch}{0.9}

\resizebox{\linewidth}{!}{%
\setlength{\tabcolsep}{3.0pt}
\begin{tabular}{lccccc}
\toprule
Methods
& Dual-Ser.
& DSCD
& $z_t$
& \#TP (M)
& PB\_T50\_RS \\
\midrule

\rowcolor{ultralightgray}
\multicolumn{6}{c}{\textit{Full Fine-tuning Controls}} \\
\midrule

Full fine-tuning
&  &  & -- & 16.90 & 92.05 \\

Full fine-tuning + Dual-Ser.
& \cmark &  & -- & 16.90 & 92.12 \\

Full fine-tuning + DSCD
& \cmark & \cmark & -- & 17.02 & 92.46 \\

\midrule

\rowcolor{ultralightgray}
\multicolumn{6}{c}{\textit{Conventional PEFT Controls}} \\
\midrule

Adapter
&  &  & -- & 0.87 & 84.45 \\

Adapter + DSCD
& \cmark & \cmark & -- & 0.99 & 86.08 \\

LoRA
&  &  & -- & 0.91 & 83.19 \\

LoRA ($B,C,\Delta$)
&  &  & -- & 0.91 & 89.41 \\

LoRA ($B,C,\Delta$) + DSCD
& \cmark & \cmark & -- & 1.03 & 90.63 \\

\midrule

\rowcolor{ultralightgray}
\multicolumn{6}{c}{\textit{State-Aware Adaptation}} \\
\midrule

SAA w/o block context
& \cmark &  &  & 0.72 & 89.88 \\

SAA
& \cmark &  & \cmark & 0.72 & 91.86 \\

\rowcolor{lightgreen}
\textbf{Mantis}
& \cmark & \cmark & \cmark & 0.84 & \textbf{93.48} \\

\bottomrule
\end{tabular}%
}

\end{minipage}
\hfill
% =========================================================
% Right: Few-shot learning
% =========================================================
\begin{minipage}[t]{0.47\textwidth}
\vspace{0pt}
\centering

\captionof{table}{
Few-shot learning on ModelNet40~\citep{wu20153d}.
We report the average accuracy (\%) $\pm$ standard deviation (\%) over 10 independent experiments without voting.
}
\label{tab:fewshot}

\small
\renewcommand{\arraystretch}{1.06}

\resizebox{\linewidth}{!}{%
\begin{tabular}{lcccc}
\toprule
\multirow{2}{*}{Methods}
& \multicolumn{2}{c}{5-way}
& \multicolumn{2}{c}{10-way} \\
\cmidrule(lr){2-3}
\cmidrule(lr){4-5}
& 10-shot & 20-shot & 10-shot & 20-shot \\
\midrule

\rowcolor{ultralightgray}
\multicolumn{5}{c}{\textit{Self-Supervised Learning (Full Fine-tuning)}} \\
\midrule

Point-BERT~\citep{yu2022pointbert}
& 94.6$\pm$3.1
& 96.3$\pm$2.7
& 91.0$\pm$5.4
& 92.7$\pm$5.1 \\

MaskPoint~\citep{liu2022masked}
& 95.0$\pm$3.7
& 97.2$\pm$1.7
& 91.4$\pm$4.0
& 93.4$\pm$3.5 \\

Point-MAE~\citep{pang2022pointmae}
& 96.3$\pm$2.5
& 97.8$\pm$1.8
& 92.6$\pm$4.1
& 95.0$\pm$3.0 \\

ReCon~\citep{qi2023recon}
& 97.3$\pm$1.9
& 98.9$\pm$1.2
& 93.3$\pm$3.9
& 93.3$\pm$3.9 \\

PointGPT-S~\citep{chen2023pointgpt}
& \textbf{98.0$\pm$1.9}
& \textbf{99.0$\pm$1.0}
& \textbf{94.1$\pm$3.3}
& \underline{96.1$\pm$2.8} \\

ACT~\citep{dongautoencoders}
& 96.8$\pm$2.3
& 98.0$\pm$1.4
& 93.3$\pm$4.0
& 95.6$\pm$2.8 \\

\midrule

\rowcolor{ultralightgray}
\multicolumn{5}{c}{\textit{Self-Supervised Learning (Parameter-Efficient Fine-tuning)}} \\
\midrule

PointMamba
& 96.9$\pm$2.0
& \underline{99.0$\pm$1.1}
& 93.0$\pm$4.4
& 95.6$\pm$3.2 \\

\rowcolor{lightgreen}
\quad \textbf{+ Mantis}
& \underline{97.2$\pm$1.9}
& \textbf{99.0$\pm$1.0}
& \underline{93.4$\pm$4.0}
& \textbf{96.4$\pm$3.1} \\

Mamba3D
& 96.4$\pm$2.2
& 98.2$\pm$1.2
& 92.4$\pm$4.1
& 95.2$\pm$2.9 \\

\rowcolor{lightgreen}
\quad \textbf{+ Mantis}
& 95.3$\pm$3.5
& 98.9$\pm$2.4
& 91.7$\pm$3.2
& 95.3$\pm$3.2 \\

ZigzagPointMamba
& 96.0$\pm$2.1
& 99.0$\pm$1.2
& 90.0$\pm$2.2
& 94.2$\pm$1.0 \\

\rowcolor{lightgreen}
\quad \textbf{+ Mantis}
& 96.2$\pm$1.6
& \underline{99.0$\pm$1.1}
& 92.8$\pm$2.1
& 95.5$\pm$4.6 \\

\bottomrule
\end{tabular}%
}

\end{minipage}

\end{table*}

\subsection{Few-shot Learning}
We further conduct few-shot experiments on ModelNet40\citep{wu20153d} to assess our few-shot transfer learning ability. Following previous works\citep{pang2022pointmae}, we employ the n-way, m-shot configuration with $n \in \{5,10\}$ and $m \in \{10,20\}$. As shown in Table~\ref{tab:fewshot}, our Mantis achieves the best or second-best accuracy in all configurations, indicating its effectiveness in few-shot learning. This consistent trend across configurations highlights Mantis's ability to reduce performance variance, which is crucial for robust few-shot learning under limited data.

\subsection{Detailed results for classification and part segmentation}

To provide a more fine-grained evaluation, we report per-category results for both object classification and part segmentation.
Table~\ref{tab:per_category_shapenetpart} reports category-wise IoU on ShapeNetPart\citep{shapenetpart}, while Table~\ref{tab:per_class_scanobjectnn} presents class-wise classification accuracy on ScanObjectNN\citep{uy-scanobjectnn-iccv19} (PB\_T50\_RS).
The results show that the improvement of Mantis is not concentrated on a few categories, but is generally consistent across different object and part classes.

\begin{table*}[!t]
\centering
\scriptsize
\vspace{-8pt}
\caption{
Per-category part segmentation results on ShapeNetPart\citep{shapenetpart}. 
We report instance-level mIoU, class-level mIoU, and category-wise IoU (\%).
}
\vspace{-8pt}
\label{tab:per_category_shapenetpart}
\resizebox{\textwidth}{!}{
\renewcommand{\arraystretch}{1.05}
\begin{tabular}{lcccccccccccccccccc}
\toprule
Methods 
& \rotatebox{90}{Inst.} 
& \rotatebox{90}{Cls. } 
& \rotatebox{90}{Air.} 
& \rotatebox{90}{Bag} 
& \rotatebox{90}{Cap} 
& \rotatebox{90}{Car} 
& \rotatebox{90}{Chair} 
& \rotatebox{90}{Ear.} 
& \rotatebox{90}{Guitar} 
& \rotatebox{90}{Knife} 
& \rotatebox{90}{Lamp} 
& \rotatebox{90}{Laptop} 
& \rotatebox{90}{Motor.} 
& \rotatebox{90}{Mug} 
& \rotatebox{90}{Pistol} 
& \rotatebox{90}{Rocket} 
& \rotatebox{90}{Skate.} 
& \rotatebox{90}{Table} \\
\midrule
\rowcolor{ultralightgray}
\multicolumn{19}{c}{\textit{Full Fine-tuning}} \\
\midrule
PointMamba
& 85.3 & 82.6
& 83.3 & 83.5 & 86.6 & 79.0 & 90.0 & 77.5 & 90.7 & 86.1 
& 83.9 & 94.6 & 72.8 & 93.4 & 83.5 & 60.4 & 74.7 & 81.6 \\

Mamba3D
& 85.6 & 83.6
& 83.7 & 84.3 & 87.5 & 76.5 & 91.2 & 80.6 & 91.2 & 87.5 
& 87.1 & 95.6 & 72.2 & 94.2 & 83.7 & 66.5 & 72.5 & 83.4 \\

ZigzagPointMamba
& 85.8 & 84.2
& 84.3 & 84.7 & 87.9 & 78.5 & 91.2 & 80.6 & 91.2 & 87.5 
& 87.1 & 95.6 & 74.2 & 94.2 & 83.7 & 68.6 & 74.6 & 83.3 \\

\midrule
\rowcolor{ultralightgray}
\multicolumn{19}{c}{\textit{Parameter-Efficient Fine-tuning}} \\
\midrule
\rowcolor{lightgreen}
PointMamba + Mantis
& 85.1 & 82.9
& 83.0 & 84.0 & 87.0 & 78.6 & 89.8 & 78.5 & 90.6 & 86.5 
& 84.9 & 94.8 & 73.3 & 93.6 & 83.7 & 62.2 & 74.8 & 81.1 \\

\rowcolor{lightgreen}
Mamba3D + Mantis
& 86.1 & 84.1
& 83.9 & 84.7 & 87.9 & 78.2 & 91.2 & 80.4 & 91.0 & 87.5 
& 87.1 & 95.6 & 74.2 & 94.2& 83.7 & 68.1 & 74.6 & 83.1 \\

\rowcolor{lightgreen}
ZigzagPointMamba + Mantis
& 85.9 & 84.5
& 84.5 & 84.9 & 88.1 & 78.3 & 91.4 & 81.1 & 91.3 & 87.8 
& 87.7 & 95.8 & 74.5 & 94.4 & 83.9 & 70.1 & 74.6 & 83.6 \\

\bottomrule
\end{tabular}
}
\vspace{-8pt}
\end{table*}

\begin{table*}[!t]
\centering
\scriptsize

\caption{
Per-class classification accuracy on ScanObjectNN\citep{uy-scanobjectnn-iccv19} (PB\_T50\_RS). We report mean class accuracy (mAcc), overall accuracy (OA), and class-wise accuracy (\%).
}
\vspace{-8pt}
\label{tab:per_class_scanobjectnn}
\resizebox{\textwidth}{!}{
\renewcommand{\arraystretch}{1.05}
\begin{tabular}{lccccccccccccccccc}
\toprule
Methods 
& \rotatebox{90}{mAcc} 
& \rotatebox{90}{OA} 
& \rotatebox{90}{Bag} 
& \rotatebox{90}{Bin} 
& \rotatebox{90}{Box} 
& \rotatebox{90}{Cab.} 
& \rotatebox{90}{Chair} 
& \rotatebox{90}{Desk} 
& \rotatebox{90}{Display} 
& \rotatebox{90}{Door} 
& \rotatebox{90}{Shelf} 
& \rotatebox{90}{Table} 
& \rotatebox{90}{Bed} 
& \rotatebox{90}{Pillow} 
& \rotatebox{90}{Sink} 
& \rotatebox{90}{Sofa} 
& \rotatebox{90}{Toilet} \\
\midrule
\rowcolor{ultralightgray}
\multicolumn{18}{c}{\textit{Full Fine-tuning}} \\
\midrule
PointMamba
& 87.6 & 89.3 
& 81.0 & 89.2 & 75.2 & 89.1 & 92.8 & 85.5 & 89.0 & 92.4
& 88.9 & 87.6 & 86.8 & 88.9 & 87.5 & 92.0 & 87.7 \\

Mamba3D 
& 90.2 & 92.1 
& 83.8 & 92.0 & 78.2 & 91.6 & 95.5 & 88.2 & 91.4 & 95.2 
& 91.6 & 90.2 & 89.4 & 91.3 & 89.8 & 94.6 & 90.2 \\

ZigzagPointMamba
& 86.9 & 88.7 
& 79.9 & 88.9 & 73.6 & 88.5 & 92.3 & 84.5 & 88.7 & 92.3 
& 88.5 & 86.8 & 86.2 & 88.2 & 86.6 & 91.5 & 87.0 \\

\midrule
\rowcolor{ultralightgray}
\multicolumn{18}{c}{\textit{Parameter-Efficient Fine-tuning}} \\
\midrule
\rowcolor{lightgreen}
PointMamba + Mantis 
& 88.3 & 90.0 
& 82.5 & 90.0 & 77.0 & 89.6 & 92.9 & 86.8 & 89.4 & 92.6 
& 89.3 & 88.6 & 87.9 & 89.3 & 88.3 & 92.2 & 88.1 \\

\rowcolor{lightgreen}
Mamba3D + Mantis 
& 91.6 & 93.5 
& 86.5 & 93.1 & 81.5 & 92.7 & 95.8 & 90.2 & 92.6 & 95.6 
& 92.5 & 91.7 & 91.2 & 92.5 & 91.6 & 95.1 & 91.4 \\

\rowcolor{lightgreen}
ZigzagPointMamba + Mantis 
& 88.5 & 90.1 
& 82.9 & 90.0 & 77.7 & 89.6 & 93.0 & 87.1 & 89.8 & 92.8 
& 89.6 & 88.6 & 88.0 & 89.4 & 88.5 & 92.2 & 88.3 \\
\bottomrule
\end{tabular}
}
\end{table*}

For part segmentation, Mantis maintains competitive instance-level mIoU and improves class-level mIoU across all three Mamba-based backbones.
The gains on categories such as “Bag”, “Cap”, “Knife”, “Lamp”, “Mug”, and “Rocket” suggest that Mantis is effective in preserving fine-grained local geometric structures under parameter-efficient adaptation.
For object classification, Mantis consistently improves both mAcc and OA over the corresponding full fine-tuning baselines, with clear gains on challenging categories such as “Bag”, “Box”, “Desk”, “Bed”, “Pillow”, and “Sink”.
These categories often exhibit larger intra-class variation or stronger geometric ambiguity, indicating that state-aware adaptation and cross-serialization consistency help improve robustness under noisy and order-sensitive real-world point cloud inputs.

\subsection{Part segmentation visualization}
In Figure~\ref{fig:part_seg1}, \ref{fig:part_seg3}, and \ref{fig:part_seg2}, we visualize our Mantis part segmentation results on the Mamba3D\citep{li2024mamba3d} baseline. We present all 16 ShapeNetPart\citep{shapenetpart} categories from four different viewpoints. Mantis accurately segments large, well-defined parts such as airplane wings and chair seats, while also capturing small or complex parts like guitar strings and mug handles. The four-view visualizations demonstrate robustness to camera angles and occlusions, maintaining consistent part predictions. Even fine-grained structures are well-preserved, highlighting effective feature aggregation and state-aware adaptation. Color-coded part labels separate adjacent parts, reducing ambiguity. Despite achieving these visual results, Mantis introduces only a small number of trainable parameters, confirming its parameter-efficient design. Overall, these visualizations validate that Mantis provides accurate, robust, and efficient fine-grained part segmentation across diverse objects.

\subsection{Additional Results on Large-scale Semantic Segmentation}
\label{app:efficiency}

\begin{table*}[!t]
\centering

\caption{
Semantic segmentation results on the ScanNet V2 validation set\citep{dai2017scannet}.
We report voxel-level mean IoU (VmIoU), point-level mean IoU (PmIoU),
point-level per-class IoU (\%), and trainable backbone parameters (M).
Among all PEFT methods, the best and second-best results are highlighted in black and underlined, respectively.
}

\label{tab:scannet}

\fontsize{4.2}{5.3}\selectfont
\setlength{\tabcolsep}{0.5pt}
\renewcommand{\arraystretch}{1.05}

\resizebox{\textwidth}{!}{%
\begin{tabular}{l|ccc|*{20}{c}}
\toprule

\multirow[c]{2}{*}{%
  \raisebox{2.8ex}[0pt][0pt]{Method}%
} &

\multicolumn{3}{c|}{%
  \raisebox{4.0ex}[0pt][0pt]{ScanNet V2 Val~\cite{dai2017scannet}}%
} &

\classlabel{wall} &
\classlabel{floor} &
\classlabel{cabinet} &
\classlabel{bed} &
\classlabel{chair} &
\classlabel{sofa} &
\classlabel{table} &
\classlabel{door} &
\classlabel{window} &
\classlabel{books.} &
\classlabel{picture} &
\classlabel{counter} &
\classlabel{desk} &
\classlabel{curtain} &
\classlabel{refrig.} &
\classlabel{shower} &
\classlabel{toilet} &
\classlabel{sink} &
\classlabel{bathtub} &
\classlabel{other}
\\[-2.8ex]

\noalign{%
  \global\advance\aboverulesep by -0.5ex
  \global\advance\belowrulesep by 0.5ex
}
\cmidrule{2-4}
\noalign{%
  \global\advance\aboverulesep by 0.5ex
  \global\advance\belowrulesep by -0.5ex
}

&
\raisebox{1ex}[0pt][0pt]{Param.} &
\raisebox{1ex}[0pt][0pt]{VmIoU} &
\raisebox{1ex}[0pt][0pt]{PmIoU} &

\classcolor{wallcolor} &
\classcolor{floorcolor} &
\classcolor{cabinetcolor} &
\classcolor{bedcolor} &
\classcolor{chaircolor} &
\classcolor{sofacolor} &
\classcolor{tablecolor} &
\classcolor{doorcolor} &
\classcolor{windowcolor} &
\classcolor{bookshelfcolor} &
\classcolor{picturecolor} &
\classcolor{countercolor} &
\classcolor{deskcolor} &
\classcolor{curtaincolor} &
\classcolor{refrigeratorcolor} &
\classcolor{showercolor} &
\classcolor{toiletcolor} &
\classcolor{sinkcolor} &
\classcolor{bathtubcolor} &
\classcolor{othercolor}
\\

% =========================================================
% Full Fine-tuning
% =========================================================

\specialrule{\lightrulewidth}{0.4ex}{0.45ex}

\rowcolor{sectiongray}
\multicolumn{24}{c}{%
\fontsize{4.3}{4.7}\selectfont\itshape
Self-Supervised Learning (Full Fine-tuning)
} \\

\specialrule{\lightrulewidth}{0.45ex}{0.4ex}

Point-MAE~\citep{pang2022pointmae}
& 21.82
& \other{51.2}
& \other{50.8}
& \other{70.8}
& \other{91.1}
& \other{47.5}
& \other{63.8}
& \other{66.0}
& \other{55.4}
& \other{52.3}
& \other{38.1}
& \other{35.8}
& \other{47.8}
& \other{28.0}
& \other{33.4}
& \other{34.6}
& \other{46.8}
& \other{51.7}
& \other{41.3}
& \other{73.1}
& \other{48.5}
& \other{59.1}
& \other{30.9}
\\

ReCon~\citep{qi2023recon}
& 21.82
& \other{50.8}
& \other{50.5}
& \other{70.4}
& \other{90.4}
& \other{43.3}
& \other{64.5}
& \other{65.0}
& \other{53.8}
& \other{52.2}
& \other{37.2}
& \other{36.0}
& \other{46.9}
& \other{27.9}
& \other{33.5}
& \other{32.2}
& \other{44.7}
& \other{54.2}
& \other{43.2}
& \other{75.3}
& \other{49.4}
& \other{57.5}
& \other{32.4}
\\

Mamba3D~\citep{li2024mamba3d}
& 15.39 (100\%)
& \other{51.3}
& \other{50.7}
& \other{69.8}
& \other{87.6}
& \other{43.2}
& \other{62.7}
& \other{67.3}
& \other{54.4}
& \other{54.7}
& \other{40.2}
& \other{35.7}
& \other{45.9}
& \other{29.4}
& \other{34.5}
& \other{38.0}
& \other{50.0}
& \other{54.8}
& \other{41.1}
& \other{71.0}
& \other{45.6}
& \other{57.8}
& \other{30.3}
\\

% =========================================================
% General PEFT Methods
% =========================================================

\specialrule{\lightrulewidth}{0.4ex}{0.45ex}

\rowcolor{sectiongray}
\multicolumn{24}{c}{%
\fontsize{4.3}{4.7}\selectfont\itshape
General\hspace{2pt}PEFT\hspace{2pt}Methods
} \\

\specialrule{\lightrulewidth}{0.45ex}{0.4ex}

Mamba3D (lin.)
& 0.00 (0.00\%)
& \other{40.3}
& \other{39.8}
& \other{60.6}
& \other{79.5}
& \other{34.8}
& \other{50.0}
& \other{55.6}
& \other{45.6}
& \other{41.5}
& \other{29.0}
& \other{26.5}
& \other{35.5}
& \other{16.9}
& \other{22.8}
& \other{23.1}
& \other{36.7}
& \other{41.4}
& \other{32.2}
& \other{61.0}
& \other{36.6}
& \other{47.2}
& \other{19.5}
\\

+ Adapter~\citep{houlsby2019parameter}
& 1.90 (12.35\%)
& \other{39.9} (\drop{-0.4})
& \other{39.5} (\drop{-0.3})
& \other{60.4}
& \other{76.2}
& \other{34.1}
& \other{48.9}
& \other{55.2}
& \other{45.4}
& \other{41.8}
& \other{28.8}
& \other{26.2}
& \other{33.4}
& \other{19.0}
& \other{22.0}
& \other{23.7}
& \other{36.6}
& \other{43.2}
& \other{30.2}
& \other{62.1}
& \other{35.1}
& \other{49.3}
& \other{18.4}
\\

+ Prefix Tuning~\citep{li2021prefix}
& 1.67 (10.85\%)
& \other{40.8} (\gain{+0.5})
& \other{40.4} (\gain{+0.6})
& \second{63.9}
& \other{74.2}
& \other{36.8}
& \other{51.8}
& \other{53.8}
& \other{45.0}
& \other{40.4}
& \other{28.7}
& \other{25.7}
& \second{41.6}
& \other{19.9}
& \other{21.7}
& \other{24.1}
& \other{36.0}
& \other{41.8}
& \other{32.4}
& \other{62.4}
& \other{36.8}
& \other{50.0}
& \other{21.0}
\\

+ LoRA~\citep{hu2022lora}
& 1.98 (12.87\%)
& \other{42.7} (\gain{+2.4})
& \other{42.2} (\gain{+2.4})
& \other{63.6}
& \other{82.4}
& \second{39.8}
& \second{57.3}
& \other{56.1}
& \other{49.1}
& \other{45.9}
& \other{32.3}
& \other{30.6}
& \other{41.3}
& \other{24.2}
& \second{27.4}
& \other{24.1}
& \second{43.1}
& \other{42.1}
& \best{43.0}
& \other{47.7}
& \second{44.0}
& \other{27.1}
& \other{22.9}
\\

% =========================================================
% PEFT Methods for Point Cloud
% =========================================================

\specialrule{\lightrulewidth}{0.4ex}{0.45ex}

\rowcolor{sectiongray}
\multicolumn{24}{c}{%
\fontsize{4.3}{4.7}\selectfont\itshape
PEFT\hspace{2pt}Methods\hspace{2pt}for\hspace{2pt}Point\hspace{2pt}Cloud
} \\

\specialrule{\lightrulewidth}{0.45ex}{0.4ex}

+ IDPT~\citep{zha2023idpt}
& 2.11 (13.71\%)
& \other{39.6} (\drop{-0.7})
& \other{39.1} (\drop{-0.7})
& \other{57.0}
& \other{75.5}
& \other{33.7}
& \other{53.0}
& \other{56.3}
& \other{45.5}
& \other{40.9}
& \other{26.2}
& \other{27.7}
& \other{37.4}
& \other{15.8}
& \other{18.7}
& \other{21.7}
& \other{39.8}
& \other{42.3}
& \other{28.8}
& \other{60.2}
& \other{35.1}
& \other{47.7}
& \other{18.7}
\\

+ DAPT~\citep{zhou2024dapt}
& 1.93 (12.54\%)
& \other{41.5} (\gain{+1.2})
& \other{41.0} (\gain{+1.2})
& \other{62.0}
& \other{80.7}
& \other{31.4}
& \other{51.0}
& \other{54.9}
& \other{47.8}
& \other{43.4}
& \other{30.7}
& \other{26.9}
& \other{37.0}
& \other{25.2}
& \other{20.5}
& \other{23.3}
& \other{38.2}
& \other{41.1}
& \other{33.5}
& \other{61.7}
& \other{40.8}
& \other{48.9}
& \other{21.0}
\\

+ PMA~\citep{zha2025pma}
& 1.77 (11.50\%)
& \other{43.1} (\gain{+2.8})
& \other{42.3} (\gain{+2.5})
& \other{61.3}
& \other{81.5}
& \other{36.3}
& \other{55.2}
& \second{56.9}
& \other{47.2}
& \second{46.5}
& \second{32.5}
& \other{28.1}
& \other{35.8}
& \other{20.3}
& \other{24.9}
& \second{27.9}
& \other{38.0}
& \other{43.3}
& \other{34.9}
& \other{61.6}
& \other{40.2}
& \other{50.3}
& \second{23.3}
\\

+ PointGST~\citep{liang2024pointgst}
& 1.05 (6.82\%)
& \other{43.2} (\gain{+2.9})
& \other{42.7} (\gain{+2.9})
& \other{62.6}
& \other{83.0}
& \other{34.9}
& \other{55.2}
& \other{55.3}
& \other{48.6}
& \other{42.3}
& \other{31.4}
& \second{34.7}
& \other{35.2}
& \best{27.8}
& \other{23.4}
& \other{26.8}
& \other{36.7}
& \second{46.1}
& \other{33.7}
& \second{64.0}
& \other{40.6}
& \second{54.4}
& \other{17.3}
\\

+ PoinTriE~\citep{sun2026tri}
& 1.08 (7.02\%)
& \second{43.9} (\gain{+3.6})
& \second{43.4} (\gain{+3.6})
& \other{62.5}
& \best{89.0}
& \other{35.9}
& \other{55.8}
& \other{56.5}
& \second{49.8}
& \other{44.7}
& \other{32.0}
& \other{31.7}
& \other{36.5}
& \other{21.8}
& \other{23.9}
& \other{27.4}
& \other{38.4}
& \other{45.5}
& \other{37.7}
& \other{63.1}
& \other{39.9}
& \other{54.1}
& \other{21.8}
\\

\rowcolor{lightgreen}
+ Mantis
& 0.73 (4.74\%)
& \best{50.4} (\gain{+10.1})
& \best{50.0} (\gain{+10.2})
& \best{70.3}
& \second{88.4}
& \best{44.5}
& \best{62.3}
& \best{64.4}
& \best{55.1}
& \best{52.6}
& \best{38.3}
& \best{36.2}
& \best{45.4}
& \second{27.1}
& \best{32.2}
& \best{34.5}
& \best{47.3}
& \best{52.4}
& \second{41.5}
& \best{71.2}
& \best{47.4}
& \best{58.7}
& \best{30.2}
\\

\noalign{\vskip -0.5ex}
\bottomrule

\end{tabular}%
}

\end{table*}

Table~\ref{tab:scannet} presents detailed results on ScanNet V2\citep{dai2017scannet}.
Mantis achieves 50.4 VmIoU and 50.0 PmIoU with 0.73M trainable backbone parameters.
Compared with existing PEFT methods, Mantis shows a consistent performance advantage while using a relatively small number of trainable parameters, and its performance remains close to full fine-tuning.
The class-wise results further show that the gains are distributed across diverse semantic categories rather than concentrated on a few classes, reflecting the stable adaptation capability of Mantis across heterogeneous scene semantics.

\begin{table*}[!t]
\centering

\caption{Efficiency comparison on ScanNet V2\citep{dai2017scannet}.}

\label{tab:efficiency_scannet}

\scriptsize
\setlength{\tabcolsep}{3pt}
\renewcommand{\arraystretch}{0.95}

\resizebox{\textwidth}{!}{%
\begin{tabular}{l|cc|cccccc|cc}
\toprule

Method
& \shortstack{Train.\\Backbone (M)}
& \shortstack{Head\\(M)}
& \shortstack{Inf. FLOPs\\(G)}
& \shortstack{Peak Mem.\\(GB)}
& \shortstack{AdamW States\\(GB)}
& \shortstack{Train Time\\(min/ep.)}
& \shortstack{Latency\\(ms/scene)}
& \shortstack{Storage\\(MB)}
& \shortstack{VmIoU\\(\%)}
& \shortstack{PmIoU\\(\%)}
\\

\midrule

\textbf{Point-MAE}
& 21.82
& 5.20
& 64.8
& 19.2
& 0.216
& 19.4
& 91.0
& 108.1
& 51.2
& 50.8
\\

\quad + PointLoRA (CVPR'25)
& 2.19
& 5.20
& 65.7
& 18.5
& 0.059
& 18.2
& 94.0
& 29.6
& 38.1
& 37.2
\\

\quad + PMA (CVPR'25)
& 2.35
& 5.20
& 67.9
& 18.8
& 0.060
& 19.0
& 99.5
& 30.2
& 39.3
& 37.9
\\

\quad + PoinTriE (ECCV'26)
& 1.90
& 5.20
& 68.7
& 18.8
& 0.056
& 19.3
& 100.6
& 27.6
& 43.4
& 42.1
\\

\midrule

\textbf{PointMamba}
& 12.15
& 6.13
& 41.5
& 14.8
& 0.146
& 14.5
& 64.5
& 73.1
& 50.3
& 50.1
\\

\quad + PMA (CVPR'25)
& 1.17
& 6.13
& 46.8
& 14.4
& 0.058
& 16.7
& 73.2
& 29.2
& 42.7
& 41.6
\\

\rowcolor{lightgreen}
\quad + \textbf{Mantis (Ours)}
& \textbf{0.60}
& \textbf{6.13}
& \textbf{43.4}
& \textbf{13.8}
& \textbf{0.055}
& \textbf{15.3}
& \textbf{66.1}
& \textbf{27.5}
& \textbf{49.8}
& \textbf{49.5}
\\

\midrule

\textbf{Mamba3D}
& 15.39
& 6.47
& 48.7
& 16.5
& 0.175
& 16.3
& 72.8
& 87.4
& 51.3
& 50.7
\\

\quad + PMA (CVPR'25)
& 1.77
& 6.47
& 54.8
& 16.0
& 0.066
& 18.6
& 82.4
& 33.0
& 43.1
& 42.3
\\

\rowcolor{lightgreen}
\quad + \textbf{Mantis (Ours)}
& \textbf{0.73}
& \textbf{6.47}
& \textbf{51.1}
& \textbf{15.2}
& \textbf{0.058}
& \textbf{16.9}
& \textbf{75.0}
& \textbf{28.8}
& \textbf{50.4}
& \textbf{50.0}
\\

\bottomrule
\end{tabular}%
}

\vspace{-14pt}
\end{table*}

Table~\ref{tab:efficiency_scannet} provides a detailed efficiency comparison on ScanNet v2.
On PointMamba/Mamba3D, Mantis reduces trainable backbone parameters by
95.1\%/95.3\%, peak training memory by 6.8\%/7.9\%,
AdamW-state memory by 62.3\%/66.9\%, and per-task storage by
62.4\%/67.0\%, respectively.
These substantial efficiency gains come with limited computational overhead,
requiring only 1.9/2.4G additional FLOPs, 0.8/0.6 min per epoch,
and 1.6/2.2 ms per scene.
Meanwhile, Mantis maintains performance close to full fine-tuning
and consistently improves over PMA, indicating a favorable overall balance
between practical efficiency and downstream performance.

\section{Explanatory experiments and discussions}

\subsection{Training dynamics and optimization stability}

\begin{figure}[!tbp] 
  \centering
  \includegraphics[width=1\linewidth]{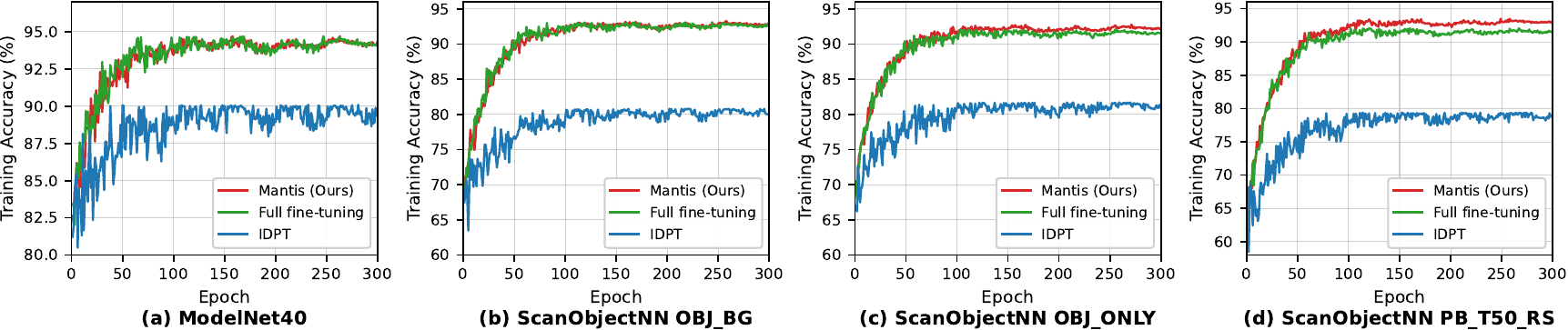} 
  \vspace{-18pt}
  \caption{The classification accuracy curves of full fine-tuning, IDPT\citep{zha2023idpt}, and our Mantis with Mamba3D on ModelNet40\citep{wu20153d} and ScanObjectNN\citep{uy-scanobjectnn-iccv19}.} 
  \label{fig:training_dynamics}

\end{figure}

\begin{figure}[htbp] 
  \centering
  \includegraphics[width=1\linewidth,height=0.8\textheight]{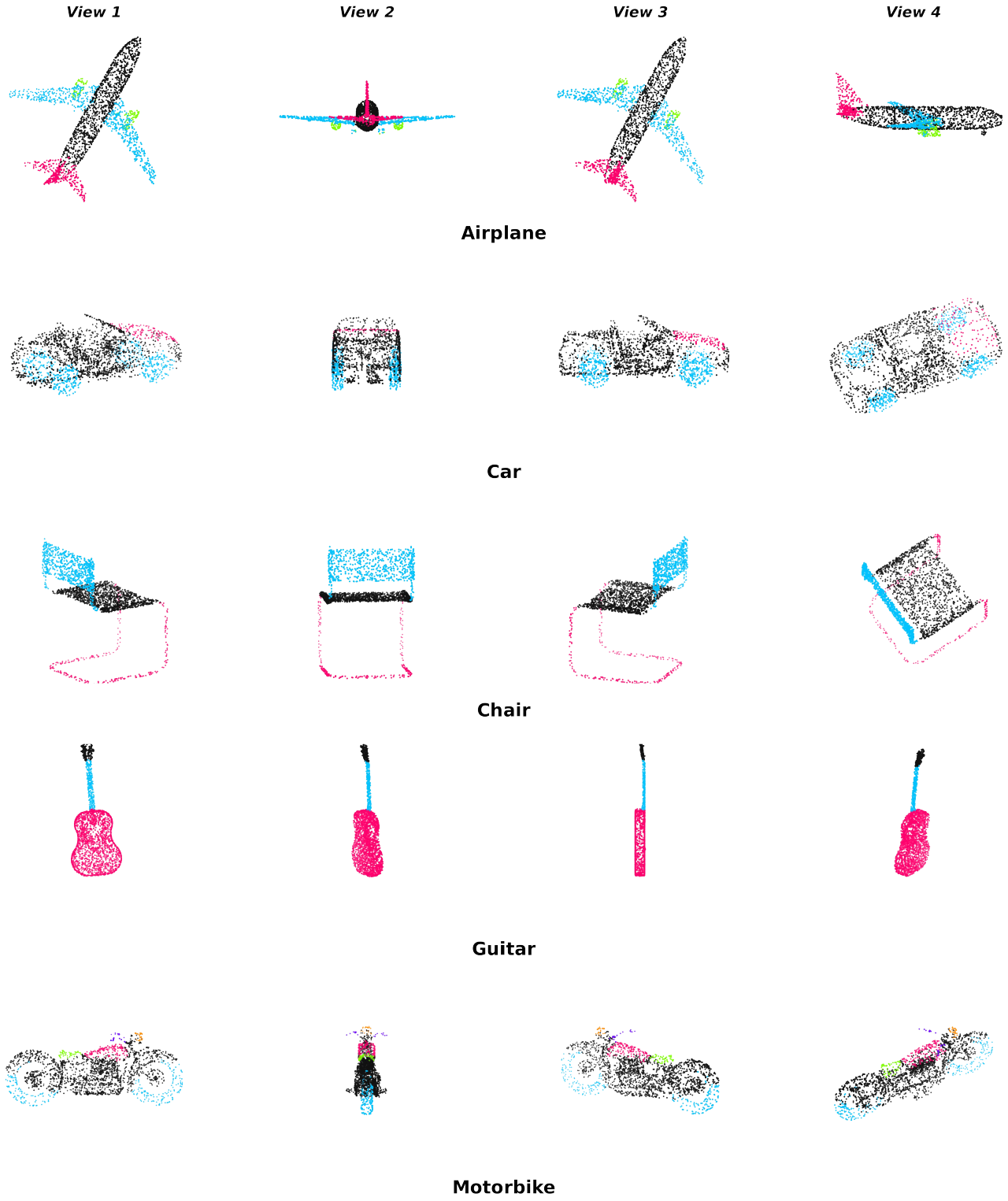} 
  \caption{\textbf{Visualization results for part segmentation on ShapeNetPart\citep{shapenetpart}}. Projected prediction images from Mantis are shown across four different viewpoints, including the categories “Airplane”, “Car”, “Chair”, “Guitar” and “Motorbike”.}
  \label{fig:part_seg1} 
\end{figure}

\begin{figure}[htbp] 
  \centering
  \includegraphics[width=1\linewidth]{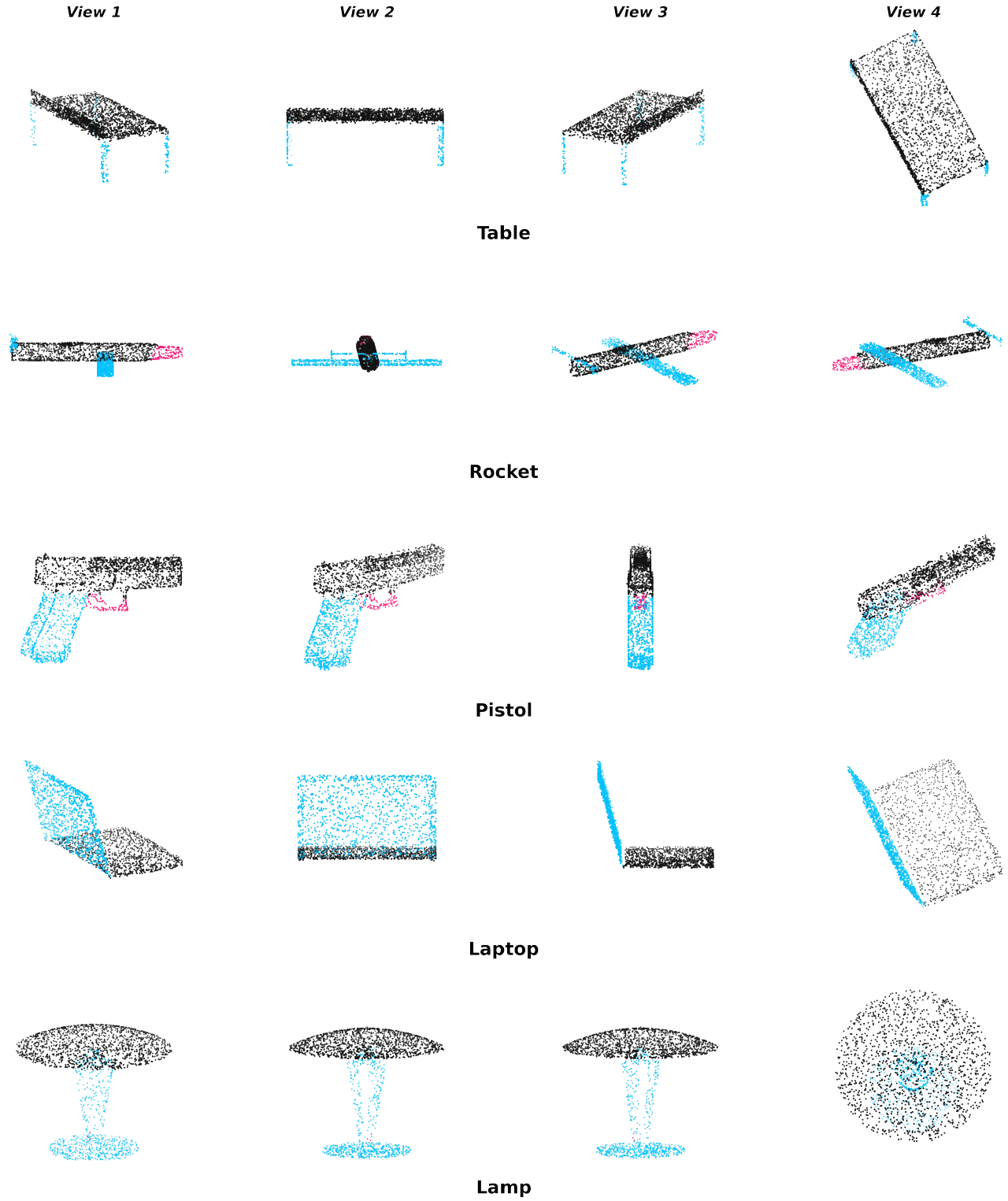} 
  \caption{\textbf{Visualization results for part segmentation on ShapeNetPart\citep{shapenetpart}}. Projected prediction images from Mantis are shown across four different viewpoints, including the categories “Table”, “Rocket”, “Pistol”, “Laptop” and “Lamp”.}
  \label{fig:part_seg3} 
\end{figure}

\begin{figure}[htbp] 
  \centering
  \includegraphics[width=1\linewidth]{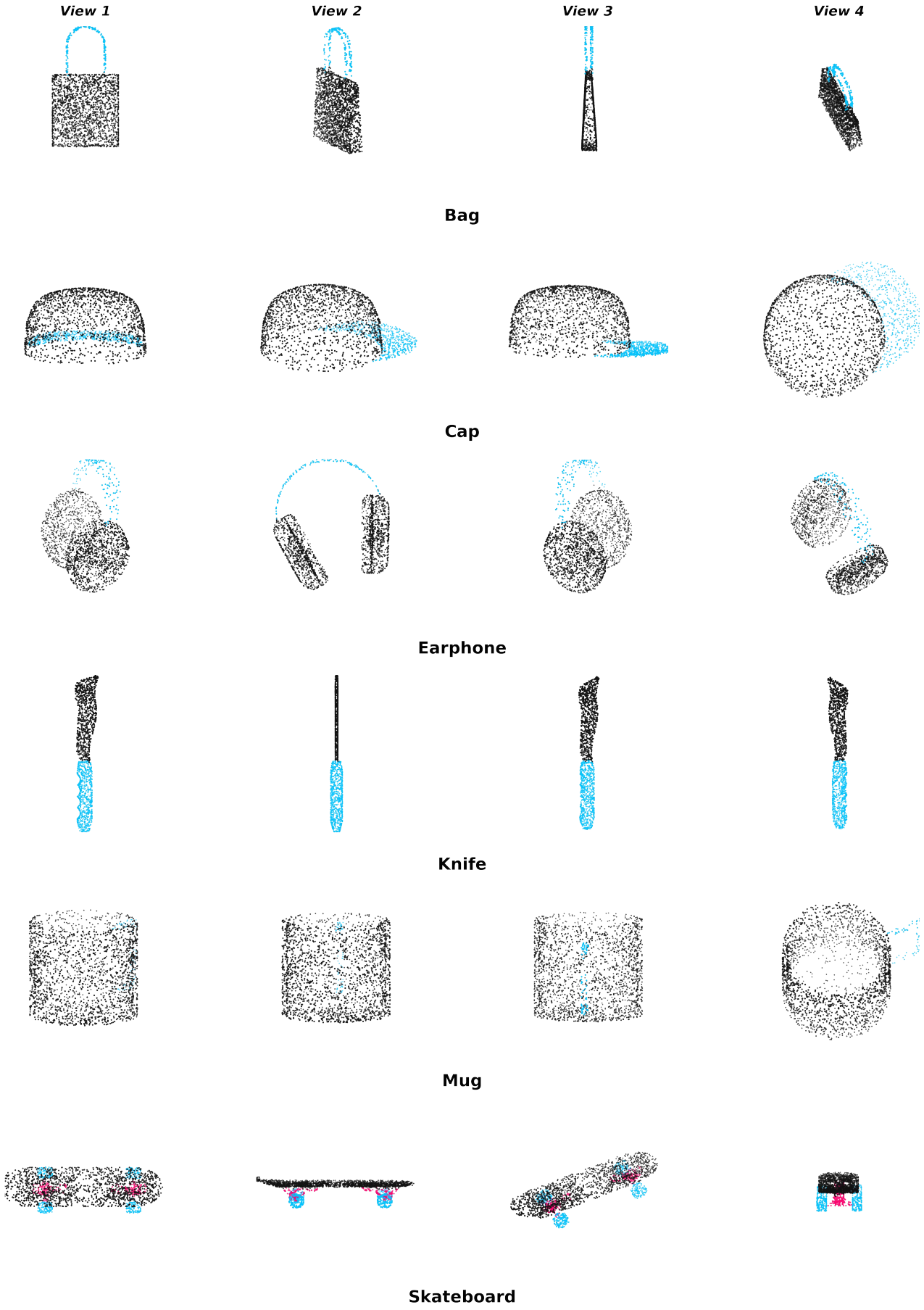} 
  \caption{\textbf{Visualization results for part segmentation on ShapeNetPart\citep{shapenetpart}}. Projected prediction images from Mantis are shown across four different viewpoints, including the categories “Bag”, “Cap”, “Earphone”, “Knife”, “Mug” and “Skateboard”.}
  \label{fig:part_seg2} 
\end{figure}

\begin{insightbox}{Observation: Token-Level Prompting Destabilizes Mamba Adaptation}
Directly applying IDPT\citep{zha2023idpt} to Mamba3D\citep{li2024mamba3d} leads to unstable training and inferior convergence, suggesting that token-level prompt tuning is poorly aligned with Mamba's state-level sequence dynamics.

\begin{itemize}
    \item \textbf{Mismatch:} IDPT is designed for Transformer-style token interactions, while Mamba relies on selective state-space dynamics and recurrent hidden-state propagation.
    
    \item \textbf{Mantis Design:} SAA adapts the state evolution directly, and DSCD regularizes predictions across different serialization orders.
    
    \item \textbf{Empirical Evidence:} Figure~\ref{fig:training_dynamics} illustrates that IDPT exhibits unstable training and slower convergence, whereas Mantis stabilizes training across challenging variants.
\end{itemize}
Mantis achieves smoother and more stable convergence, reaching performance comparable to or better than full fine-tuning with only about 5\% of trainable parameters.
\end{insightbox}
Although IDPT\citep{zha2023idpt} is effective for Transformer-based point cloud backbones, directly applying it to Mamba3D\citep{li2024mamba3d} leads to unstable training dynamics and inferior convergence on downstream classification tasks. 
This observation is consistent with our motivation that token-level prompt tuning is not well aligned with the state-level sequence dynamics of Mamba. 
In contrast, Mantis introduces state-aware adaptation and dual-serialization consistency, which better match the selective state-space modeling mechanism. 
As a consequence, Mantis shows smoother and more stable convergence, especially on the challenging ScanObjectNN variants, and achieves final performance comparable to or better than full fine-tuning with only about 5\% of trainable parameters.

\vspace{-8pt}
\subsection{Further analysis on PB\_T50\_RS}
\label{sec:app_hardest}

An interesting phenomenon in Table~\ref{tab:classification} is that the advantage of Mantis becomes more pronounced on the most challenging PB\_T50\_RS split than on the relatively easier OBJ\_BG and OBJ\_ONLY variants. We conjecture that this trend is closely related to the order sensitivity of Mamba-style sequence modeling on point clouds. Since point clouds are inherently unordered, the hidden-state propagation of a serialized selective SSM depends on the adopted traversal order. When the input is relatively clean and structurally regular, such order sensitivity may have only a limited effect. However, under more challenging real-world scans with stronger clutter, occlusion, and geometric corruption, different valid serializations are more likely to induce larger discrepancies in state propagation and feature aggregation. In this regime, the robustness benefit of Mantis becomes more evident.

To verify this explanation, we measure the feature-level and prediction-level cross-serialization discrepancies on ScanObjectNN variants using the same formulations as the consistency terms in Sec.~\ref{sec:dscd}. We compute both quantities on the test set and report their average values without the training weights $\alpha$ and $\beta$. Lower values indicate stronger consistency across serialization views.

\begin{insightbox}{Key Insight: Why Mantis Excels on PB\_T50\_RS}
PB\_T50\_RS strongly exposes the order-sensitivity of frozen Mamba backbones in real-world point clouds. Mantis effectively addresses this through:

\begin{itemize}
    \item \textbf{State-Aware Adapter (SAA):} Dynamically modulates state-space transitions to adapt hidden states under noisy or partially corrupted inputs.
    \item \textbf{Dual-Serialization Consistency Distillation (DSCD):} Aligns representations and predictions across different serialization orders, mitigating instability.
    \item \textbf{Empirical Evidence:} Table~\ref{tab:serialization_discrepancy} shows that Mantis consistently reduces feature-level and prediction-level cross-serialization discrepancies, with the largest reduction on PB\_T50\_RS.
\end{itemize}

These effects stabilize downstream predictions and yield the largest gains on PB\_T50\_RS. This does not mean PB\_T50\_RS is easier. Rather, it simply highlights weaknesses of frozen Mamba backbones under noisy, order-sensitive inputs, where Mantis can be particularly effective.
\end{insightbox}

\begin{table}[!tbp]
\centering
\caption{Cross-serialization discrepancy on different ScanObjectNN\citep{uy-scanobjectnn-iccv19} variants.}
\label{tab:serialization_discrepancy}
\resizebox{1\linewidth}{!}{
\begin{tabular}{lcccccc}
\toprule
\multirow{2}{*}{Methods} &
\multicolumn{2}{c}{OBJ\_BG} &
\multicolumn{2}{c}{OBJ\_ONLY} &
\multicolumn{2}{c}{PB\_T50\_RS} \\
\cmidrule(lr){2-3} \cmidrule(lr){4-5} \cmidrule(lr){6-7}
& Feat. Disc. & Pred. Disc. & Feat. Disc. & Pred. Disc. & Feat. Disc. & Pred. Disc. \\
\midrule
Mamba3D + Full Fine-tuning & 0.148 & 0.031 & 0.136 & 0.027 & 0.221 & 0.054 \\
\rowcolor{lightgreen}Mamba3D + Mantis & 0.109 (\textcolor{red}{$\downarrow$ 26.4\%}) & 0.019 (\textcolor{red}{$\downarrow$ 38.7\%}) & 0.101 (\textcolor{red}{$\downarrow$ 25.7\%}) & 0.017 (\textcolor{red}{$\downarrow$ 37.0\%}) & 0.128 (\textcolor{red}{$\downarrow$ 42.1\%}) & 0.024 (\textcolor{red}{$\downarrow$ 55.6\%}) \\
\bottomrule
\end{tabular}}
\end{table}

As shown in Table~\ref{tab:serialization_discrepancy}, the baseline model exhibits the largest feature-level and prediction-level discrepancy on PB\_T50\_RS among the three ScanObjectNN variants, confirming that harder real-world point clouds indeed amplify serialization sensitivity. In contrast, Mantis consistently reduces both discrepancies across all three variants, with the most substantial reduction observed on PB\_T50\_RS. This observation supports our hypothesis that the stronger performance of Mantis on the hardest split mainly stems from improved robustness to serialization-induced instability.

From a modeling perspective, this behavior can be understood from the complementary effects of the two components in Mantis. On the one hand, SAA dynamically modulates the selective state-space dynamics conditioned on the current input, incoming block context, and serialization condition, enabling the frozen Mamba backbone to adapt its state evolution more flexibly under noisy and partially corrupted observations. On the other hand, DSCD explicitly regularizes the two serialization branches by aligning their feature representations and predictive distributions, thereby mitigating the instability caused by different traversal orders. When real-world perturbation and serialization ambiguity become more severe, the advantage of Mantis becomes particularly evident.

Importantly, this phenomenon should not be interpreted as PB\_T50\_RS being easier. Rather, it indicates that PB\_T50\_RS more strongly exposes the weakness of frozen Mamba backbones under noisy and order-sensitive inputs, thus creating a setting in which our robustness-oriented design is especially effective.

\subsection{Discussion on the scope of Mamba-native adaptation}

We further clarify the scope of ``Mamba-native'' adaptation in this work. PMA\citep{zha2025pma} is an important 3D PEFT baseline that introduces Mamba into point cloud adaptation. However, its design motivation and adaptation target are different from ours. PMA is motivated by the observation that intermediate features of a frozen pre-trained model contain complementary information that is often discarded by conventional PEFT methods. Accordingly, PMA constructs an ordered sequence from intermediate-layer features and employs a Mamba adapter to fuse cross-layer semantics for downstream tasks. In its original formulation, the backbone is a frozen Transformer encoder, while Mamba is introduced as an orthogonal adapter for feature fusion. Notably, PMA focuses its evaluation on Transformer-style pre-trained backbones such as Point-BERT\citep{yu2022pointbert}, Point-MAE\citep{pang2022pointmae}, PointGPT\citep{chen2023pointgpt}, and ReCon\citep{qi2023recon}, further reflecting its original scope as a feature-fusion adapter for existing PFMs rather than a method tailored to frozen Mamba backbone dynamics.

\begin{insightbox}{Clarification: Scope of Mamba-Native Adaptation}
While PMA\citep{zha2025pma} introduces Mamba modules for feature fusion on Transformer backbones, it does not adapt the frozen Mamba backbone itself. In contrast, Mantis directly modulates the native selective coefficients $(\boldsymbol{B}_t, \boldsymbol{C}_t, \boldsymbol{\Delta}_t)$ inside frozen Mamba backbones, thereby adapting state evolution without altering the pretrained structured transition operator $\boldsymbol{A}$.
\begin{itemize}
    \item \textbf{PMA:} Acts as an auxiliary adapter on intermediate features; evaluated on Transformer-style PFMs; cannot fully exploit Mamba backbone dynamics.
    \item \textbf{Mantis:} Targets the backbone itself, aligning PEFT with Mamba’s state-level dynamics, achieving stronger performance on frozen Mamba backbones.
    \item \textbf{Implication:} Designing Mamba-native PEFT requires adaptation mechanisms matched to state-space modeling, rather than reusing Mamba modules for PEFT.
\end{itemize}
Overall, Mantis is not about using Mamba as an external module, but about serving Mamba backbones through state-level adaptation.
\end{insightbox}

In contrast, Mantis targets frozen Mamba-based point cloud backbones themselves. Instead of inserting a Mamba module outside the backbone to aggregate intermediate features, Mantis adapts the selective state-space dynamics inside frozen Mamba backbones by modulating the native input-dependent coefficients $(\boldsymbol{B}_t, \boldsymbol{C}_t, \boldsymbol{\Delta}_t)$ while retaining the structured transition operator $\boldsymbol{A}$ frozen. Therefore, we consider ``Mamba-native 3D PEFT'' not as the first use of Mamba modules in 3D PEFT, but as a state-level adaptation framework specifically tailored to frozen Mamba-based point cloud backbones. This distinction can be summarized as follows: \textbf{PMA uses Mamba as an auxiliary adapter, whereas Mantis is designed for adapting Mamba backbones.}

This distinction also motivates our choice of PMA as a representative PEFT baseline. Since PMA is the closest existing 3D PEFT method involving Mamba modules, comparing with it allows us to examine whether a general Mamba-based adapter designed for feature fusion can be directly transferred to frozen Mamba backbones. The results show that such feature-level adapter fusion is not sufficient for Mamba-backbone adaptation. In contrast, Mantis consistently achieves stronger performance by directly modulating the internal state-space operators. This comparison highlights the necessity of designing PEFT mechanisms that are aligned with the state-level dynamics of Mamba, rather than only reusing Mamba as an external feature-fusion module.

\section{Symbol table}

\begin{table*}[ht]
\centering
\small
\vspace{-18pt}
\caption{Symbol table}
\vspace{-6pt}
\label{tab:symbols}
\renewcommand{\arraystretch}{1.02}
\begin{tabular}{p{0.32\textwidth} p{0.62\textwidth}}
\toprule
\textbf{Symbol} & \textbf{Meaning} \\
\midrule
$P \in \mathbb{R}^{M \times 3}$ & Input point cloud with $M$ points. \\
$p \in \mathbb{R}^{n \times 3}$ & Sampled key points after Farthest Point Sampling (FPS). \\
$p_h, p_{h'}$ & Two valid serialized sequences of the same point cloud. \\
$K$ & Number of nearest neighbors in each local patch. \\
$d, d_h$ & Token feature dimension and hidden-state dimension. \\
$d_o$ & Dimension of learnable order embedding. \\
$d_\phi$ & Dimension of the shared control space in SAA. \\
$\phi_t^{(k)}$ & Fused control representation at step $t$ for branch $k$. \\
$E_h^{(1)}, E_{h'}^{(2)}$ & Token features under the two serialization orders. \\
$E^{(k)}$ & Token features of branch $k$ ($k\in\{1,2\}$). \\
$o^{(k)}$ & Learnable order embedding for branch $k$. \\
$e^{(k)}$ & Order-aware representation for branch $k$. \\
$\mathcal{G}(\cdot)$ & Lightweight transformation for order-aware representation. \\
$\tilde{Z}_l^{(k)}$ & LN-projected feature in $l$-th encoder block. \\
$X_l^{(k)}$ & Intermediate feature fed into selective SSM. \\
$Z_l^{(k)}$ & Output of $l$-th encoder block. \\
$Z_0^{(k)}, Z_L^{(k)}$ & Input/final feature of SAA-Mamba encoder. \\
$z^{(k)}$ & Projected global embedding in DSCD ($g_{\varphi}$). \\
$r$ & Bottleneck dimension in SAA. \\
$\alpha, \beta, \tau$ & Feature/prediction loss weights and DSCD temperature. \\
$\mathbf{q}_t^{(k)}$ & Driving vector before soft-thresholding. \\
$\boldsymbol{\lambda}_t^{(k)}$ & Adaptive threshold vector. \\
$\mathbf{u}_t^{(k)}$ & Sparse task-conditioned control signal. \\
$d_O^{\mathrm{out}}, d_O^{\mathrm{in}}$
& Output and input dimensions of selective coefficient $O$. \\
$\rho$
& Contraction bound of the frozen state transition operator. \\
$h_t^{(k,0)}$ & Hidden state from frozen backbone. \\
$\delta h_t^{(k)}$ & Hidden-state deviation under SAA. \\
$\varepsilon_{\mathrm{tr}}, \varepsilon_B$
& Bounds on discretized-transition and input-operator perturbations. \\
$H, X$ & Upper bounds on hidden-state and input norms. \\
$\varphi(\cdot)$ & Order modulation mapping. \\
$\Phi(\cdot)$ & Control-feature fusion function in SAA. \\
$\mathcal{F}_{\mathrm{SAA}}(\cdot)$ & Selective SSM enhanced with SAA. \\
$g_{\varphi}(\cdot)$ & Branch-shared projection head in DSCD. \\
$\boldsymbol{W}_{t,i}^{(k)}$ & Selective transfer kernel from step $i$ to $t$. \\
$\boldsymbol{W}_{t,i}^{(k,0)}$ & Frozen backbone kernel without SAA. \\
$\Delta \boldsymbol{W}_{t,i}^{(k)}$ & Perturbation of selective transfer kernel. \\
$\mathbf{U}_O, \mathbf{V}_O$
& Coefficient-specific low-rank modulation matrices. \\
$\boldsymbol{\Gamma}_{O,t}^{(k)}$
& Sparse low-rank factorized correction before native-layout mapping. \\
$\delta_{O,t}^{(k)}$
& Native-layout correction applied to selective coefficient $O$. \\
$\mathcal{R}_O(\cdot)$
& Mapping from the factorized correction to the native coefficient layout. \\
$\mathcal{N}_h^{(1)}, \mathcal{N}_{h'}^{(2)}$ & Local neighborhoods under two serialization branches. \\
$k$ & Branch index for dual serialization. \\
$x_t^{(k)}, h_t^{(k)}, y_t^{(k)}$ & Input, hidden state, and output at step $t$. \\
$\ell^{(k)}, \pi^{(k)}$ & Logits and softened prediction of branch $k$. \\
$\overline{\boldsymbol{A}}, \overline{\boldsymbol{B}}$ & Discretized SSM operators (ZOH). \\
$\widehat{\boldsymbol{A}}_t^{(k)}, \widehat{\boldsymbol{B}}_t^{(k)}$
& SAA-induced discretized transition and input operators. \\
$\boldsymbol{A}$
& Frozen structured continuous-time transition operator. \\
$\boldsymbol{B}_t^{(k)}, \boldsymbol{C}_t^{(k)}, \boldsymbol{\Delta}_t^{(k)}$
& Input-dependent selective coefficients at step $t$. \\
$z_t^{(k)}$
& Incoming pre-scan block representation. \\
$\widetilde{\boldsymbol{B}}_t^{(k)}, \widetilde{\boldsymbol{C}}_t^{(k)}, \widetilde{\boldsymbol{\Delta}}_t^{(k)}$
& Selective coefficients after SAA modulation. \\
$W_x^{\phi}, W_z^{\phi}, W_e^{\phi}, W_{\mathrm{drv}}^{\psi}, W_{\mathrm{gt}}^{\psi}$
& Learnable projection matrices in the pre-scan SAA controller. \\
$\mathcal{L}_{\mathrm{task}}, \mathcal{L}_{\mathrm{feat}}, \mathcal{L}_{\mathrm{pred}}$ & Task loss, feature consistency, and prediction consistency loss. \\
\bottomrule
\end{tabular}
\end{table*}

\clearpage

\end{document}